\def\year{2021}\relax
\documentclass[letterpaper]{article} 
\usepackage{aaai21}  
\usepackage{times}  
\usepackage{helvet} 
\usepackage{courier}  
\usepackage[hyphens]{url}  
\usepackage{graphicx} 
\usepackage{natbib}  
\usepackage{caption} 
\usepackage{amsmath,amssymb} 
\usepackage{epsfig}
\usepackage{graphicx}
\usepackage{mathabx}
\usepackage{booktabs}
\usepackage{subfig}
\usepackage{appendix}

\newcommand{\real}{\mathbb{R}}

\newcommand{\figref}[1]{Fig.~\ref{#1}}
\newcommand{\tabref}[1]{Table~\ref{#1}}
\newcommand{\secref}[1]{Sec.~\ref{#1}}
\newcommand{\eqnref}[1]{Eqn.~\ref{#1}}

\title{Stereopagnosia: Fooling Stereo Networks with Adversarial Perturbations}
\author {
    Alex Wong$^\dagger$\thanks{ $\dagger$ denotes authors with equal contributions.}, \
    Mukund Mundhra$^\dagger$, \
    Stefano Soatto \\
}
\affiliations {
    UCLA Vision Lab \\
    alexw@cs.ucla.edu, mukundmundhra@cs.ucla.edu, soatto@cs.ucla.edu
}

\renewcommand\footnotemark{}
\begin{document}

\maketitle

\begin{abstract}
We study the effect of adversarial perturbations of images on the estimates of disparity by deep learning models trained for stereo. We show that imperceptible additive perturbations can significantly alter the disparity map, and correspondingly the perceived geometry of the scene. These perturbations not only affect the specific model they are crafted for, but transfer to models with different architecture, trained with different loss functions. We show that, when used for adversarial data augmentation, our perturbations result in trained models that are more robust, without sacrificing overall accuracy of the model. This is unlike what has been observed in image classification, where adding the perturbed images to the training set makes the model less vulnerable to adversarial perturbations, but to the detriment of overall accuracy. We test our method using the most recent stereo networks and evaluate their performance on public benchmark datasets.
\end{abstract}

\section{Introduction}
\label{sec:introduction}

Deep Neural Networks are seen as fragile, in the sense that small perturbations of their input, for instance an image, can cause a large change in the output, for instance the inferred class of objects in the scene \cite{moosavi2016deepfool} or its depth map \cite{wong2020targeted}. This is not too surprising, since there are infinitely many scenes that are consistent with the given image, so at inference time one has to rely on the complex relation between that image and {\em different scenes} portrayed in the training set. 
This is not the case for stereo: Under mild assumptions discussed below, a depth map can be uniquely inferred point-wise from two images. There is no need to {\em learn} stereo, as the images of a particular scene are sufficient to infer its depth without relying on images of different scenes. (The reason we {\em do} use learning is to regularize the reconstruction where the assumptions mentioned below are violated, for instance in regions of homogeneous reflectance.) It would therefore be surprising if one could perturb the images in a way that forces the model to over-rule the evidence and alter the perceived depth map, especially if such perturbations affect regions of non-uniform reflectance. In this paper, we show that this {\em can} be done and refer to this phenomenon as \textit{stereopagnosia}, a geometric analogue of prosopagnosia \cite{damasio1982prosopagnosia}.

Specifically, we consider stereo networks, that are functions that take as input a calibrated stereo pair and produce a depth map as output. A stereo pair consists of two images captured by cameras in known relative configuration (position and orientation), with non-zero parallax (distance between the optical centers), projectively rectified so that corresponding points (points in the two image planes whose pre-image under perspective projections intersect in space) lie on corresponding scan-lines. A depth map is a function that associates to each pixel in a rectified image a positive real number corresponding to the distance of the point of first intersection in the scene from the optical center. 

Equivalently, the network can output {\em disparity}, the displacement between corresponding points in the two images, from which depth can be computed in closed form. Wherever a point in the scene is supported on a surface that is Lambertian,  locally smooth, seen under constant illumination, co-visible from both images, and sporting a {\em sufficiently exciting} reflectance,\footnote{There exist region statistics that exhibit isolated extrema, so the region around the point is ``distinctive'' \cite{lowe1999object}.} its distance from the images can be computed in closed-form \cite{ma2012invitation}. Where such assumptions are violated, disparity is either not defined, for instance in occluded regions that are visible from one image but not the other, or ill-posed, for instance in regions with constant reflectance where any disparity is equally valid (the so-called ``aperture problem''). To impute disparity to these regions, regularization can be either generic ({\em e.g.}, minimal-surface assumptions \cite{horn1981determining}) or data-driven, exploiting known relations between stereo pairs and disparity in scenes other than the one in question. This is where stereo networks come in.

Our first contribution is to show that {\em stereo networks are vulnerable to adversarial perturbations,} which are small additive changes in the input images (either one or both), designed for a specific image pair in a way that maximally changes the output of a specific trained deep network model. The fact that it is possible to alter the disparity, {\em even in regions that satisfy the assumptions discussed above} (\figref{fig:ifgsm-mifgsm}), where disparity is uniquely defined and computable in closed form, is surprising since the network is forced to ignore the evidence, rather than simply exploit the unbounded hypothesis space available in an ill-posed problem. 

The second contribution is to show that, despite being crafted for a specific model, the perturbations can affect the behavior of other models, with different network architecture, trained with different loss functions and optimization methods. However, transferability is not symmetric, for instance perturbations constructed for AANet \cite{xu2020aanet} can wreak havoc if used with DeepPruner \cite{duggal2019deeppruner}, but not vice-versa. Models that incorporate explicit matching, such as correlation, are more robust than those that are agnostic to the mechanics of correspondence, and are instead based on stacking generic features. 

Our third contribution is more constructive, and establishes that adversarial perturbations can be used to beneficial effects by augmenting the dataset and function as regularizers. Unlike in single-image classification and monocular depth perception where such regularization trades off robustness to perturbations with overall accuracy, in our case we obtain models that are more robust while retaining the performance of the original model. 

To achieve these results, we extend the Fast Gradient Sign Method \cite{goodfellow2014explaining} and its iterative versions \cite{dong2018boosting, kurakin2016adversarial}, developed for for single frame classification,  to two-frame stereo disparity estimation. We evaluate the robustness of recent stereo methods (PSMNet, DeepPruner, AANet) on the standard benchmark stereo datasets (KITTI 2012 \cite{geiger2012we} and 2015 \cite{menze2015object}).


\section{Related Works}
\label{sec:related-works}
\textbf{Adversarial Perturbations} 
\cite{szegedy2013intriguing}  have been extensively studied for classification \cite{goodfellow2014explaining, szegedy2013intriguing} with many iterative methods to boost the effectiveness of the attacks \cite{dong2018boosting, kurakin2016adversarial, moosavi2016deepfool}. \cite{moosavi2017universal} further extended the attacks to the universal setting, where the same perturbations can be added to each image in a dataset to fool a network;  \cite{nguyen2015deep} showed that unrecognizable noise can result in high confidence predictions. To defend against such attacks, \cite{kurakin2016adversarial, tramer2017ensemble} proposed training with adversarial data augmentation and \cite{xie2017mitigating} improved it with randomization.

Recently, \cite{naseer2019cross} studied transferability of perturbations across datasets and models and \cite{xie2019improving} improved transferability across networks by deforming the image. \cite{peck2017lower} demonstrated lower bounds on the magnitudes of perturbations needed to fool a network and \cite{ilyas2019adversarial} showed that the existence of adversarial perturbations can be attributed to non-robust
features.

While there are many adversarial works on classification, there
exist only a few for dense-pixel prediction tasks (e.g. semantic segmentation, depth, optical flow). \cite{xie2017adversarial} designed attacks for detection and segmentation.
\cite{hendrik2017universal} demonstrated targeted universal attacks for semantic segmentation, where the network is fooled to predict a specific target. \cite{wong2020targeted} used targeted attacks to provide explainability for single image depth prediction networks; whereas \cite{dijk2019neural} probed them by inserting vehicles into input images. \cite{mopuri2018generalizable} examined universal attacks for segmentation and single image depth. \cite{ranjan2019attacking} studied patch attacks for optical flow.

Unlike \cite{mopuri2018generalizable, wong2020targeted}, we study \textit{non-targeted} adversarial perturbations for stereo matching. 
While \cite{ranjan2019attacking} also use multiple frames, they apply the \textit{same visible} patch to the same locations in both images, whereas our attacks are \textit{visually imperceptible} and crafted separately for each image. 

\textbf{Deep Stereo Matching} \cite{zagoruyko2015learning, zbontar2016stereo} leveraged deep networks to extract features and perform matching separately. Recent works implement the entire stereo pipeline as network layers trained end-to-end. \cite{mayer2016large} used correlation layers to create a 2D cost volume. \cite{pang2017cascade} extended \cite{mayer2016large} to a cascade residual learning framework. AANet \cite{xu2020aanet} also used correlation, but instead introduced adaptive sampling to replace convolutions 
when performing cost aggregation to avoid sampling at discontinuities. \cite{kendall2017end} proposed to concatenate features together to build a 3D cost volume for performing cost aggregation. PSMNet \cite{chang2018pyramid} added spatial pyramid pooling layers and introduced a stacked hourglass architecture. DeepPruner \cite{duggal2019deeppruner} followed the 3D cost volume architectures proposed by \cite{chang2018pyramid,kendall2017end} and proposed differentiable patch matching over deep features to construct a sparse 3D cost volume. 

In this work, we consider adversaries for PSMNet \cite{chang2018pyramid}, DeepPruner \cite{duggal2019deeppruner} and AANet \cite{xu2020aanet}. PSMNet is an exemplar of modern stereo networks (stacked hourglass, cost volume, 3D convolutions), but uses feature stacking without explicit matching. DeepPruner shares the general architecture of PSMNet, but performs explicit matching. AANet is the state of the art and represents the 2D convolution and correlation architecture. In choosing these methods, we (i) examine their individual robustness against adversaries (\secref{sec:attacks}), (ii) study the transferability of perturbations between similar and different architectures (\secref{sec:transferability}), and (iii) apply defenses to increase robustness against adversaries (\secref{sec:defenses}). To the best of our knowledge, we are the first to study adversarial perturbations for stereo. As mentioned in the introduction, it is not a given that adversarial perturbations, known to exist for single image reconstruction, would exist for stereo, where the geometry of the scene is uniquely determined from the data, at least in the regions that satisfy the unique correspondence assumptions.

\begin{figure*}[t]
    \centering
        \includegraphics[width=1.01\textwidth]{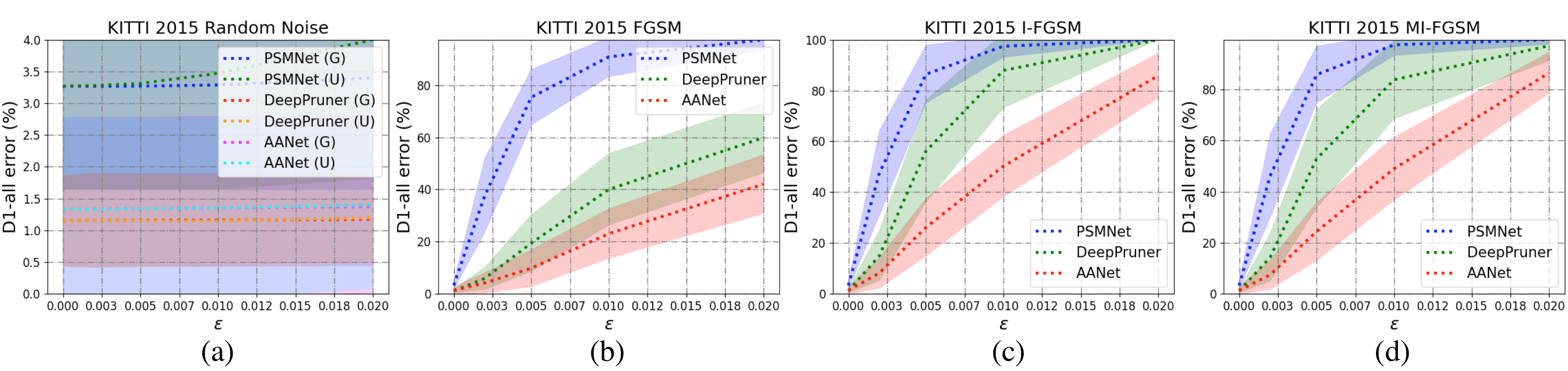}
    \caption{\textit{Attacks on stereo models.} (a) Gaussian (G) and uniform (U) noise with various upper norms ($\epsilon$) are added to input images as a naive attack. All methods are robust to both as performances across $\epsilon$ are approximately constant. (b) Even with a single optimization step, FGSM with $\epsilon=0.02$ is able increase error of PSMNet from $\approx$3\% to $\approx$97\%. (c, d) Iterative methods (I-FGSM, MI-FGSM) are able to further degrade performance, increasing error of AANet from $\approx$1\% to $\approx$87\% and as much as 100\%  for PSMNet and DeepPruner. AANet is consistently more robust to adversarial noise than PSMNet and DeepPruner.}
    \label{fig:plot-attacks}
\end{figure*}

\begin{figure}
    \centering
    \includegraphics[width=0.48\textwidth]{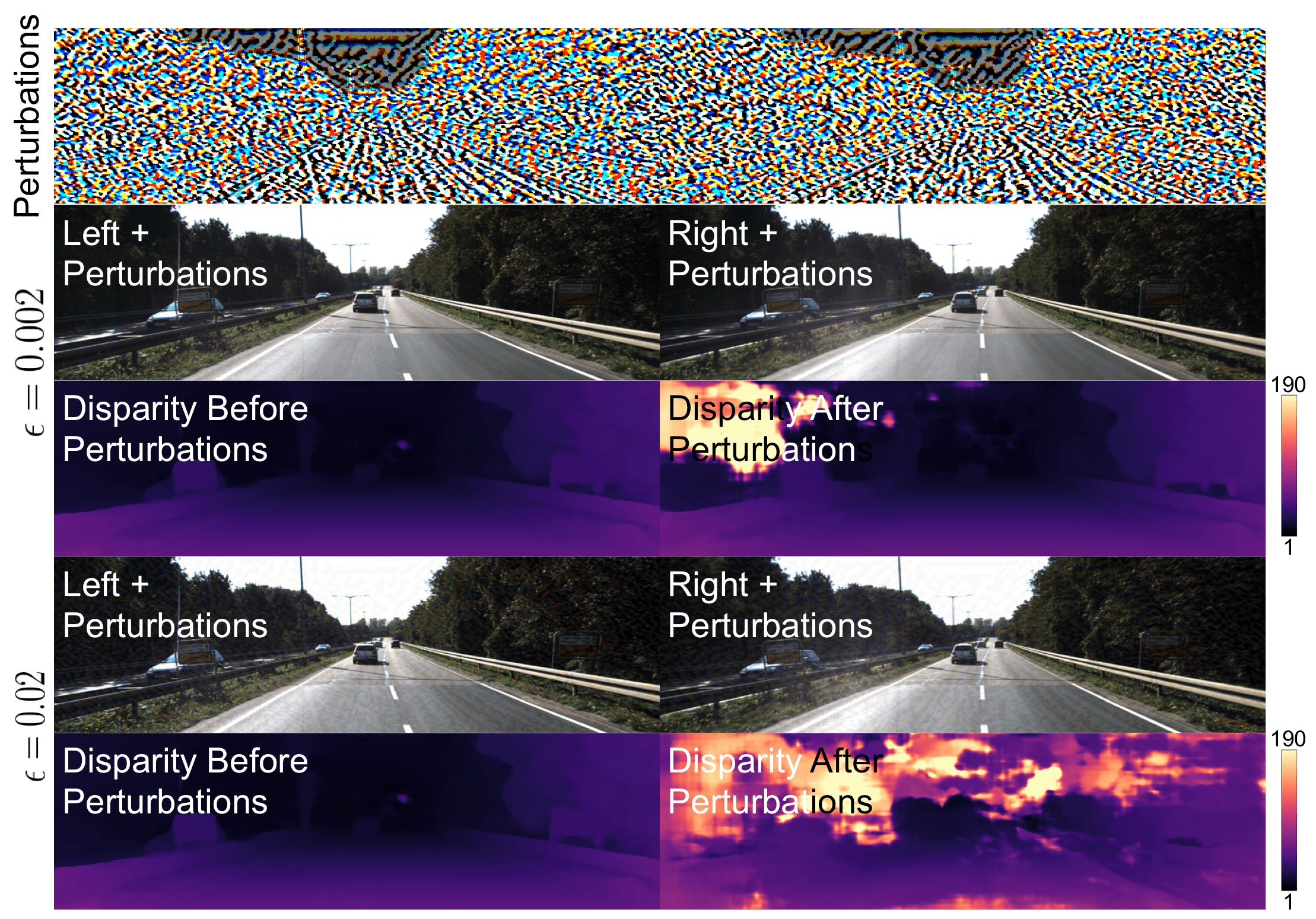}
    \caption{\textit{FGSM with upper norms of 0.002 and 0.02}. With visually imperceptible noise, $\epsilon = 0.002$, PSMNet is fooled to predict much larger disparities (closer depths) in the top left corner region. Using $\epsilon = 0.02$, the perturbations corrupt the geometry of the entire scene.}
    \label{fig:fgsm-2e2-2e3}
\end{figure}

\section{Generating Adversarial Perturbations}
\label{sec:generating-adversarial-perturbations}
Given a pretrained stereo network $f_\theta(x_L, x_R)$ that predicts the disparity between the left $x_L$ and right $x_R$ images of a stereo pair, our goal is to craft perturbations $v_L, v_R \in \real^{H \times W \times 3}$, such that when added to $(x_L, x_R)$, $f_\theta(x_L, x_R)~\neq~f_\theta(x_L+ v_L, x_R+v_R)$. To ensure that the perturbations are visually imperceptible, we subject them to the norm constraints $\|v_I\|_{\infty} \leq \epsilon$ for $I \in \{L, R\}$. To demonstrate such perturbations exist, we extend white-box methods Fast Gradient Sign Method (FGSM) \cite{goodfellow2014explaining}, iterative-FGSM (I-FGSM) \cite{kurakin2016adversarial}, and its momentum variant (MI-FGSM) \cite{dong2018boosting}, originally for classification, to the stereo matching task. We note that it is also possible to perturb only one of the two images (e.g. let $v_L = 0$ or $v_R = 0$); the effect is less pronounced, but nonetheless present and shown in the  Supp. Mat.   

\textbf{FGSM.} Assuming access to the network $f_\theta$ and its loss function $\ell(f_\theta(x_L, x_R), y_{gt})$, the perturbations for the left and right images are computed as the sign of gradient with respect to the images separately:
\begin{align}
    v_I &= \epsilon \cdot \texttt{sign}(\nabla_{x_I} \ell(f_\theta(x_L, x_R), y_{gt}),
    \label{eqn:fgsm}
\end{align}
where $y_{gt} \in \real^{H \times W}_+$ is the groundtruth and $I \in \{L, R\}$. 

\textbf{I-FGSM.} To craft perturbations $v_L$ and $v_R$ for the stereo pair $x_L$ and $x_R$ using iterative FGSM, we begin with $v^0_L = 0$ and $v^0_R = 0$ and accumulate the sign of gradient with respect to each image for $N$ steps:
\begin{align}
    g^{n+1}_I &= \nabla_{x_{I}} \ell(f_\theta(x_L + v^{n}_L, x_R + v^{n}_R), y_{gt}), \label{eqn:ifgsm-gradients} \\
    v^{n+1}_I &= \texttt{clip} \big{(}\alpha \cdot \texttt{sign}(g^{n+1}_I), -\epsilon, \epsilon \big{)},
    \label{eqn:i-fgsm}
\end{align}
where $n$ is the step, $\alpha$ is the step size and the $\texttt{clip}(\cdot, -\epsilon, \epsilon)$ operation sets any value less than $-\epsilon$ to $-\epsilon$ and any value larger than $\epsilon$ to $\epsilon$. The output perturbation is obtained after the $N$-th step, $v_L = v^N_L$ and $v_R = v^N_R$.

\textbf{MI-FGSM.} To leverage gradients from previous steps, we follow \cite{dong2018boosting} and replace the gradients (\eqnref{eqn:ifgsm-gradients}) with normalized gradients and a momentum term weighted by a positive scalar $\beta$ for $N$ steps:
\begin{align}
    g^{n+1}_I &= \frac{\nabla_{x_{I}} \ell(f_\theta(x_L + v^{n}_L, x_R + v^{n}_R), y_{gt})}{\|\nabla_{x_{I}} \ell(f_\theta(x_L + v^{n}_L, x_R + v^{n}_R), y_{gt})\|_1}, 
    \label{eqn:mifgsm-gradients} \\
    m^{n+1}_I &= \beta \cdot m^n_I + (1 - \beta) \cdot g^{n+1}_I, \\
    v^{n+1}_I &= \texttt{clip} \big{(}\alpha \cdot \texttt{sign}(m^{n+1}_I), -\epsilon, \epsilon \big{)},
    \label{eqn:mifgsm}
\end{align}
where $m^{0}_I = 0$, $v_L = v^N_L$, and $v_R = v^N_R$.

Besides crafting perturbations for specific models, we also study their transferability to different models. To this end, we take $(x_L + v_L, x_R + v_R)$ optimized for one model (e.g. PSMNet) and feed it as input to another (e.g. AANet). However, while iterative methods (I-FGSM, MI-FGSM) are more effective than FGSM at corrupting the target model, their perturbations are unlikely to transfer across models because they tend to overfit to the target model. To increase the transferability across models, we leverage diverse inputs \cite{xie2019improving} as data augmentation when crafting perturbations using I-FGSM and MI-FGSM.

\begin{figure*}[ht]
    \centering
    \includegraphics[width=1\textwidth]{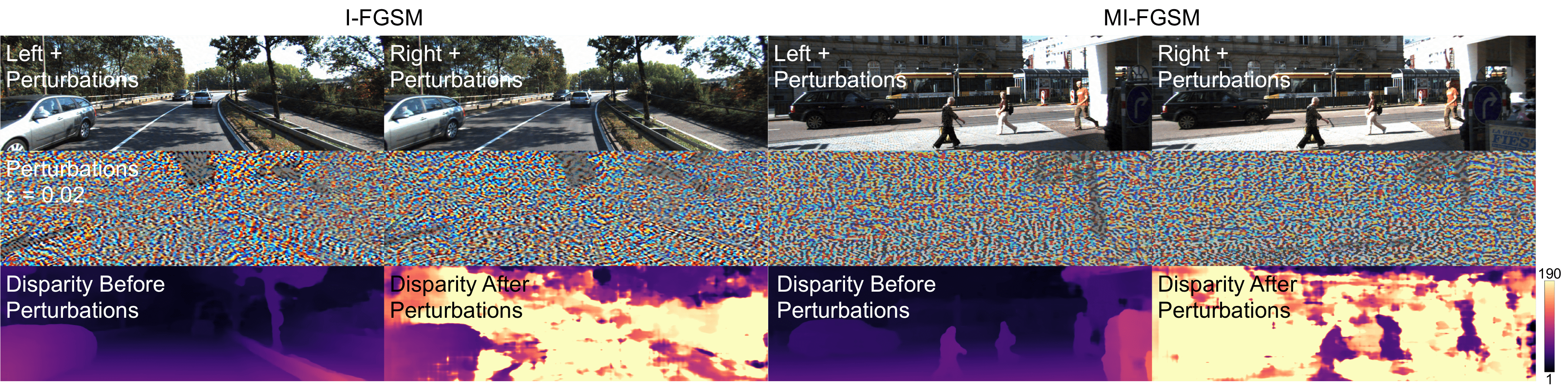}
    \caption{\textit{I-FGSM and MI-FGSM on PSMNet.} For $\epsilon = 0.02$, I-FGSM and MI-FGSM can degrade performance much more than FGSM with much smaller perturbation magnitudes. Unlike \figref{fig:fgsm-2e2-2e3} where most of the changes in disparity were concentrated on low-texture regions (no disparity signal and the perturbation drives the matching), here the perturbations degrades high texture regions. Still, the shapes of salient objects (e.g. car, human) seem to be preserved (although their disparities are altered).}
    \label{fig:ifgsm-mifgsm}
\end{figure*}

\textbf{DI$^2$-FGSM and MDI$^2$-FGSM.} Diverse inputs (DI) for iterative methods aims to reduce overfitting by randomly resizing the input images by a factor of $h \in [h_\text{min}, h_\text{max}]$ in height and $w \in [w_\text{min}, w_\text{max}]$ in width with probability $p$. To maintain the original resolution, the inputs are randomly padded with zeros on each side such that the total padding along the height is $(H - h \cdot H)$ and along the width $(W - w \cdot W)$, respectively. We denote this procedure as $\phi(x, h, w)$. However, unlike \cite{xie2019improving}, where the transformed image maps to a single class label, ground-truth disparity $y_{gt}$ is dense or semi-dense, so a matching transformation must be applied to $y_{gt}$. Moreover, the scale of $y_{gt}$ also needs to be adjusted with respect to the resized width $(w \cdot W)$ of the image. Hence, we extend diverse inputs to support stereo networks by:
\begin{align}
    \hat{x}_I &= \phi(x_I, h, w), \\
    \hat{v}_I &= \phi(v_I, h, w), \\
    \hat{y}_{gt} &= w \cdot \phi(y_{gt}, h, w)
    \label{eqn:diverse-inputs}
\end{align}
To incorporate this into iterative methods, we modify their respective gradient computations, $g_I^{n+1}$. For I-FGSM, we can re-write \eqnref{eqn:ifgsm-gradients} as:
\begin{align}
    g_I^{n+1} = \nabla_{\hat{x}_I} \ell(f_\theta(\hat{x}_L + \hat{v}^{n}_L, \hat{x}_R + \hat{v}^{n}_R), \hat{y}_{gt}),
    \label{eqn:di2fgsm}
\end{align}
Similarly, for MI-FGSM, we can modify \eqnref{eqn:mifgsm-gradients} to be:
\begin{align}
    g^{n+1}_I &= \frac{\nabla_{\hat{x}_{I}} \ell(f_\theta(\hat{x}_L + \hat{v}^{n}_L, \hat{x}_R + \hat{v}^{n}_R), \hat{y}_{gt})}{\|\nabla_{\hat{x}_{I}} \ell(f_\theta(\hat{x}_L + \hat{v}^{n}_L, \hat{x}_R + \hat{v}^{n}_R), \hat{y}_{gt})\|_1}.
    \label{eqn:mdi2fgsm}
\end{align}

To evaluate the robustness of stereo networks, we use the official KITTI D1-all (the average number of erroneous pixels in terms of disparity and end-point error) metric:
\begin{align}
    \delta(i, j) &= |f_\theta(\cdot)(i, j) - y_{gt}(i, j)|, \\
    d(i, j) &= 
    \begin{cases} 
        1 & \mbox{if } \delta(i, j) > 3, \frac{\delta(i, j)}{y_{gt}(i, j)} > 5\%, \\
        0 & \mbox{otherwise} 
    \end{cases} \\
    \text{D1-all} &= \frac{1}{\|\Omega_{gt}\|} \sum_{i, j \in \Omega_{gt}} d(i, j),
    \label{eqn:d1-all}
\end{align}
where $\Omega_{gt}$ is a subset of the image space $\Omega$ with valid ground-truth disparity annotations, $y_{gt} > 0$. 

\section{Experiment Setup}
\label{sec:experiment-setup}
\textbf{Datasets.} We evaluate adversarial perturbations (robustness, transferability, defense) for recent stereo methods (PSMNet, DeepPruner, AANet) on the standard benchmark datasets: KITTI 2015 stereo \cite{menze2015object} validation set in the main paper and KITTI 2012 \cite{geiger2012we} validation set in the Supp. Mat.

KITTI 2015 is comprised of 200 training stereo pairs and KITTI 2012 consists of 194 (all at $376 \times 1240$ resolution) with ground-truth disparities obtained using LiDAR for outdoor driving scenes. Following KITTI validation protocol, the KITTI 2015 training set is divided into 160 for training and 40 for validation, and the KITTI 2012 training set is split into 160 for training and 34 for validation. Due to computational limitations, we downsampled all images to $256 \times 640$; hence, there are slight increases in errors (\eqnref{eqn:d1-all}) compared to those reported by baseline methods. 

\textbf{Hyper-parameters.} We study perturbations under four different upper norms, $\epsilon~=~\{0.02, 0.01, 0.005, 0.002\}$. $\epsilon = 0.002$ is where adversaries have little effect on the networks and $\epsilon = 0.02$ is the norm needed to achieve $100\%$ errors on benchmark datasets. When optimizing with I-FGSM and DI$^2$-FGSM, we used $N = 40$ and $\alpha = 1/N \cdot \epsilon$ for $\epsilon~=~ \{0.01, 0.005, 0.002\}$ and $\alpha = 0.10 \epsilon$ for $\epsilon = 0.02$.  For MI-FGSM and MDI$^2$-FGSM, $\alpha = 1/N \cdot \epsilon$ for all $\epsilon$ and chose $\beta = 0.47$ for momentum. More details on hyper-parameters and run-time can be found in Supp. Mat.

\begin{figure*}[ht]
    \centering
    \includegraphics[width=0.88\textwidth]{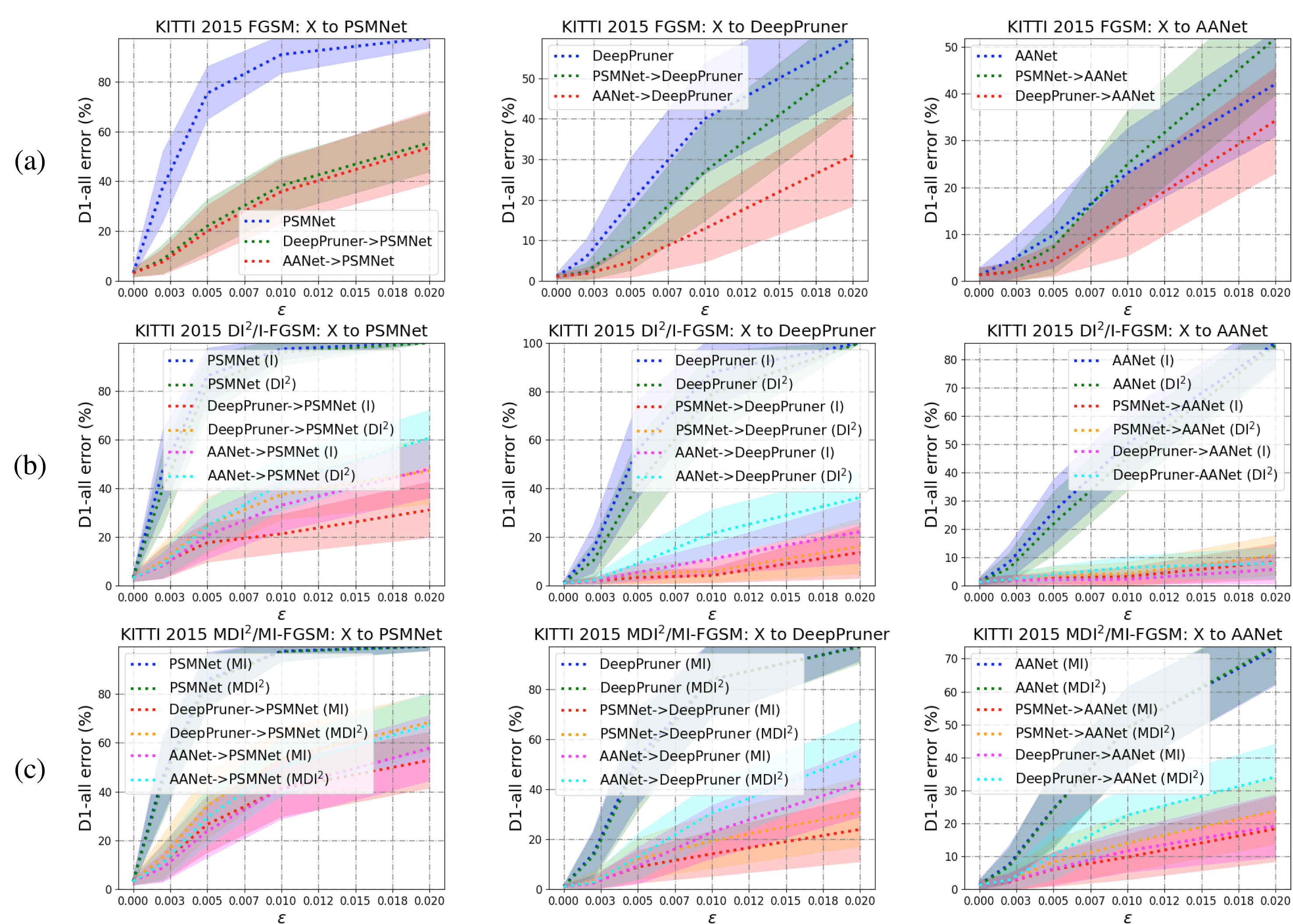}
    \caption{\textit{Transferability.} Images with added adversarial perturbation optimized for various models using (a) FGSM, (b) I-FGSM and DI$^2$-FGSM, and (c) I-FGSM and MDI$^2$-FGSM are fed as input to a target model. Transferability is not symmetric. Perturbations crafted for AANet transfer the best. AANet is also the most robust against perturbations from other models.}
    \label{fig:plot-transferability}
\end{figure*}

\section{Attacking Stereo Networks}
\label{sec:attacks}
We begin with naive attacks on stereo networks (PSMNet, DeepPruner, AANet) by perturbing the input stereo pair $(x_L, x_R)$ with Gaussian $\mathcal{N}(0, (\epsilon/4)^2)$ and uniform $\mathcal{U}(-\epsilon, \epsilon)$ noise for $\epsilon \in \{0.02, 0.01, 0.005, 0.002\}$. \figref{fig:plot-attacks}-(a) shows that such noise cannot degrade performance as the measured error stayed approximately constant under various $\epsilon$. This demonstrates that the deep features extracted for matching are robust to random noises and fooling a stereo network requires non-trivial perturbations. Hence, we examine the robustness of stereo networks against perturbations specifically optimized for each network using our variants of FGSM, I-FGSM, MI-FGSM (\secref{sec:generating-adversarial-perturbations}).

\textbf{FGSM.} \figref{fig:plot-attacks}-(a) shows errors after attacking the networks with FGSM (\eqnref{eqn:fgsm}) where perturbations are optimized over a single time step. For large upper norm $\epsilon = 0.02$, the perturbations can degrade performance significantly -- from $1.33\%$ (AANet), $1.15\%$ (DeepPruner) and $3.27\%$ (PSMNet) mean error to $42.09\%$, $59.86\%$, and $97.33\%$, respectively. The larger the upper norm, the more potent the attack, but even with small $\epsilon = 0.002$, this attack can still increase AANet to $4.18\%$ error, DeepPruner to $5.93\%$, and PSMNet to $38.11\%$. \figref{fig:fgsm-2e2-2e3} shows a comparison of FGSM attacks on PSMNet using upper norms of $0.002$ and $0.02$. For $\epsilon = 0.002$, most of the damage is localized (e.g. top left region of image space); whereas for $\epsilon = 0.02$, the entire predicted scene is corrupted. The localized damage from small norm perturbations can be attributed to the observability of the scene. We hypothesize that training affects inference where the radiance of the surfaces is not sufficiently exciting i.e. the regularizer fills in in a manner that depends on training experience. So, small norm perturbations can corrupt regions where the radiance is less informative (sky, uniform textures, foliage etc.); whereas, other regions require larger norms.

\textbf{I-FGSM, MI-FGSM.} \figref{fig:plot-attacks}-(b, c) shows that I-FGSM and MI-FGSM both affect performance similarly. Because of the multiple optimization steps, when $\epsilon = 0.02$, even the more robust AANet succumbs to the attacks -- increasing error to $\approx$87\%. For PSMNet and DeepPruner, both I-FGSM and MI-FGSM can cause them to reach 100\% error. While we cap the upper norm at $\epsilon$, we note that iterative methods are able to introduce more errors while using a much smaller perturbation magnitude. For $\epsilon = 0.02$, FGSM achieves $97.33\%$ error on PSMNet with $\|v_L\|_1 = 0.0569$ and $\|v_R\|_1 = 0.0568$; whereas, I-FGSM achieves $100\%$ error with only $\|v_L\|_1 = 0.0213$ and $\|v_R\|_1 = 0.0196$ -- less than half of the L1 norm used by FGSM, making it less perceptible. \figref{fig:ifgsm-mifgsm} shows examples of I-FGSM and MI-FGSM on PSMNet. When compared to rows 4 and 5 of \figref{fig:fgsm-2e2-2e3} ($\epsilon = 0.02$), both are less perceptible. Moreover, I-FGSM and MI-FGSM can fool the networks in textured regions where disparity can be obtained simply by matching.

\begin{figure*}[ht]
    \centering
    \includegraphics[width=1\textwidth]{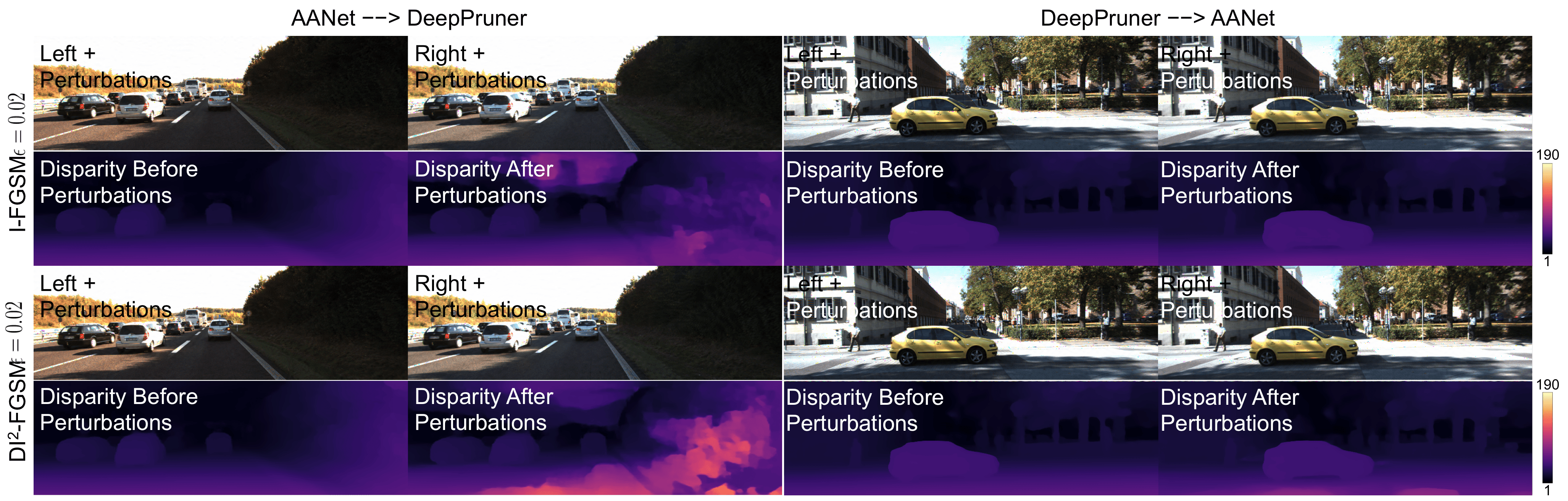}
    \caption{\textit{Transferability between AANet and DeepPruner.} We craft perturbation for AANet and DeepPruner using I-FGSM and DI$^2$-FGSM with $\epsilon = 0.02$. DI$^2$-FGSM transfers better than I-FGSM. Perturbations crafted for AANet transfer well to DeepPruner (AANet$\rightarrow$DeepPruner), but those for DeepPruner have less effect on AANet (DeepPruner$\rightarrow$AANet).}
    \label{fig:ifgsm-di2fgsm-transfer-2e2}
\end{figure*}

\textbf{Interesting observations.} Even though error reaches $100\%$ for I-FGSM and MI-FGSM, the shape (albeit incorrect) of some salient objects like cars and human still persists (\figref{fig:ifgsm-mifgsm}). This is because disparity is largely driven by the data term and hence there exist unique correspondences for such objects with sufficiently exciting texture. However, while the general shape persists, the disparity is incorrect (as it is filled in by the regularizer). Another phenomenon is that the noise required to perturb white regions (white walls, sky, \figref{fig:fgsm-2e2-2e3}, \ref{fig:ifgsm-mifgsm}) is much less than that required to attack other color intensities (e.g. uniform black). While radiance (being less informative) is a factor, we hypothesize that this is a special case due to white regions being on the upper support of image intensities, which results in high activations; hence, the adversary only needs to add small noise to adjust the activations to the needed values to corrupt the scene. We will leave the numerical analysis of this ``white-pixel'' phenomenon to future work. 

Thus, stereo networks are indeed vulnerable to adversarial perturbations. Each architecture exhibits different levels of robustness against adversaries. Feature stacking (PSMNet) is the least robust, followed by patch-matching (DeepPruner) with correlation (AANet) being the most robust. This is because DeepPruner and AANet both find correspondences based on similarity between deep features via explicit matching (well-defined data fidelity, so the perturbations corrupt the regularizer); whereas PSMNet relies purely on learned convolutional filters to produce matches and is more susceptible to the attacks.

\section{Transferability Across Models}
\label{sec:transferability}
To study transferability, we (i) optimize perturbations for PSMNet, DeepPruner and AANet separately, (ii) add each set of model-specific perturbations to the associated input stereo pair for another model i.e. add perturbations for PSMNet to the inputs of DeepPruner and AANet, and (iii) measure the resulting error using \eqnref{eqn:d1-all}. 

\textbf{FGSM, I-FGSM, MI-FGSM.} We begin with methods studied in \secref{sec:attacks}. \figref{fig:plot-transferability}-(a) shows the transferability of FGSM from different models (red, green), for which the perturbations were optimized, to a target model (blue). We found that the perturbations do transfer, but with reduced effects e.g. for $\epsilon = 0.02$, perturbations optimized for DeepPruner and AANet achieve $55.47\%$ and $53.55\%$ error on PSMNet, respectively, while perturbations optimized for PSMNet achieves $97.33\%$. The potency of the perturbations also grows with the upper norm; hence, one can increase $\epsilon$ of an adversary to further degrade new models.

\figref{fig:plot-transferability}-(b) shows the transferability of I-FGSM and \figref{fig:plot-transferability}-(c), MI-FGSM. Unlike FGSM, iterative methods transfer much less. For instance, I-FGSM ($\epsilon = 0.02$) perturbations for DeepPruner and AANet achieve $31.20\%$ and $48.08\%$, respectively, on PSMNet; whereas, FGSM achieves $55.47\%$ and $53.55\%$, respectively. In general, iterative methods transfer less than FGSM because the perturbations tend to overfit to the model they were optimized for. We note that AANet is the most robust against perturbations from other models and yet has the highest transferability, which interestingly shows that transferability is not symmetric. While perturbations for DeepPruner and AANet achieve $31.20\%$ and $48.08\%$ on PSMNet, PSMNet and AANet achieve $13.74\%$ and $22.25\%$ on DeepPruner, and those for PSMNet and DeepPruner only achieve $8.46\%$ and $5.81\%$ on AANet.  

\textbf{DI$^2$-FGSM, MDI$^2$-FGSM.} To increase transferability to other stereo networks, we additionally optimized perturbations using DI$^2$-FGSM, MDI$^2$-FGSM. \figref{fig:plot-transferability}-(b) shows that DI$^2$-FGSM (green) consistently degrades the target model's performance less than I-FGSM (blue). This is largely due to the noise in the gradients introduced by random resizing and padding. \figref{fig:plot-transferability}-(c) shows that perturbations from MDI$^2$-FGSM achieve errors similar to MI-FGSM since each iteration still retains the momentum from previous time steps.

\begin{figure*}[h]
    \centering
    \includegraphics[width=1.02\textwidth]{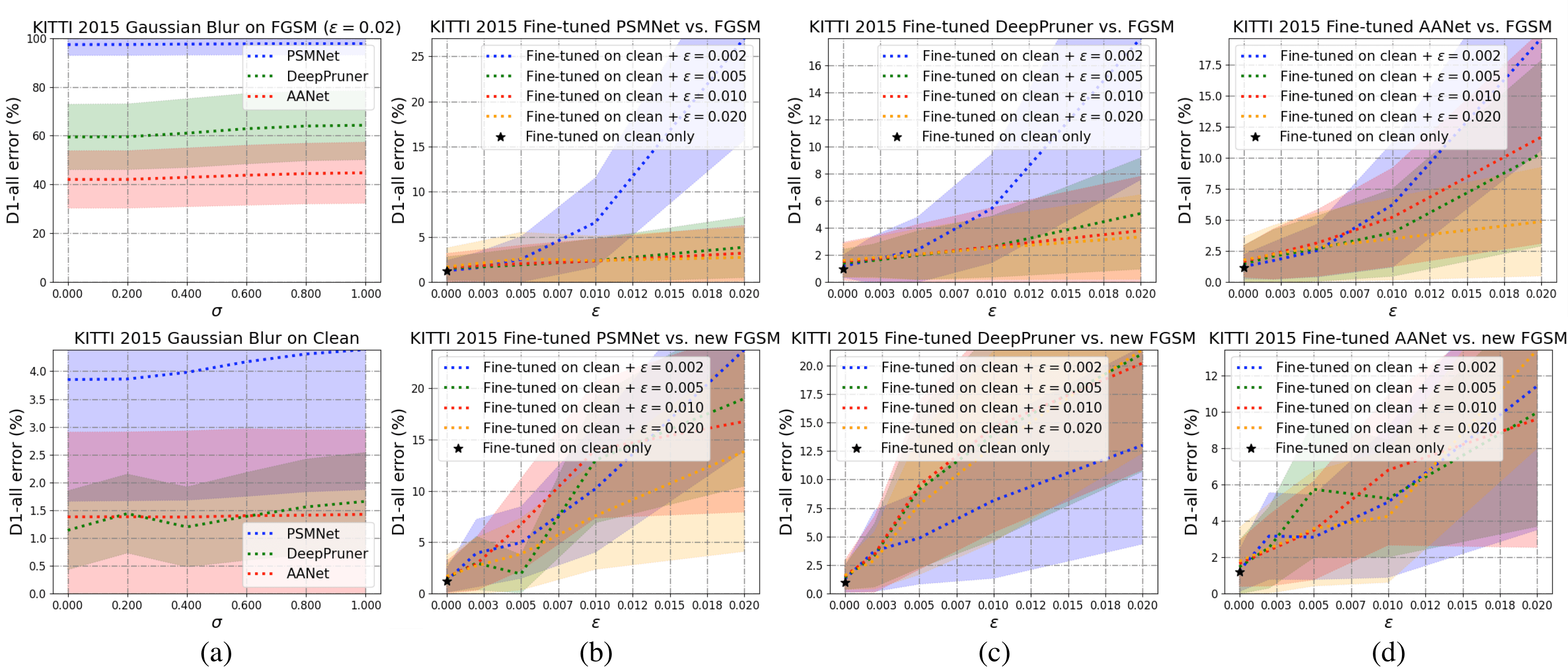}
    \caption{\textit{Defenses against attacks.} (a) Applying Gaussian blur using various $\sigma$ on perturbed (top) and clean images (bottom). Gaussian blur does not destroy perturbations, but actually degrades performance for both perturbed and clean images. Finetuning (b) PSMNet, (c) DeepPruner and (d) AANet on FGSM perturbations and defending attacks against new FGSM adversaries with various $\epsilon$. Fine-tuning on small $\epsilon$ perturbations makes the model robust against both existing and new adversaries without compromising performance on clean images.}
    \label{fig:plot-defenses}
\end{figure*}

\figref{fig:plot-transferability}-(b, c) shows that DI$^2$-FGSM and MDI$^2$-FGSM consistently transfer better to new models than I-FGSM and MI-FGSM (\figref{fig:ifgsm-di2fgsm-transfer-2e2} for visualization). The best performing iterative method is MDI$^2$-FGSM, which achieves comparable numbers to MI-FGSM on the model it is optimized for, but also transfer well to new models. We note the trends observed in I-FGSM and MI-FGSM are also observed here.  

While the mere existence of adversaries indicates (possible common) flaws in stereo networks, the reason that perturbations are transferable is because disparity is generic i.e. a surface 1m away generates the same disparity whether it belongs to a cat, a dog or a tree. Yet, transferability is not symmetric and AANet is yet again the most robust with the highest transferability. \figref{fig:ifgsm-di2fgsm-transfer-2e2} shows that perturbations optimized for AANet fools DeepPruner, but those optimized for DeepPruner have less effect on AANet. We hypothesize that architectural differences between AANet (2D convolutions) and PSMNet and DeepPruner (3D convolutions) play a role in transferability. A possible reason for why perturbations for AANet transfers better to others (but less the other way around) may be because they are optimized to attack 2D convolutional layers, which PSMNet and Deeppruner also use to build their cost volumes. However, perturbations for PSMNet and Deeppruner are optimized to disrupt 3D convolutions as well, which are not present in AANet.

\section{Defenses against Adversaries}
\label{sec:defenses}
We begin with a basic defense, Gaussian blur, against adversaries. \figref{fig:plot-defenses}-(a) shows Gaussian blurring ($3 \times 3$ kernel) with various $\sigma$ does not mitigate the effect of adversaries, but exacerbates them -- further degrading performance. In addition, simply applying Gaussian blur on clean images also decreases performance. Hence, we aim to learn more robust representations by harnessing adversarial examples to fine-tune the models. \figref{fig:plot-defenses}-(b, c, d) show the performance of stereo methods after fine-tuning on a combination of clean and perturbed images (using FGSM with various $\epsilon$). As a sanity check, we also fine-tuned on just clean images ($\star$) to ensure that differences are result of adversarial examples. 

Adversarial data augmentation increases robustness for all models. For FGSM $\epsilon = 0.02$, PSMNet decreases error from $97.33\%$ (\figref{fig:fgsm-2e2-2e3}) to $2.74\%$ against the adversary it is trained on. Moreover, training on a smaller norm ($\epsilon = 0.002$) can increase robustness against larger norm ($\epsilon = 0.02$) attacks e.g. FGSM $\epsilon = 0.02$ can degrade PSMNet to only $27.03\%$ error. Also the models are more robust against new adversaries. For this, we attacked each fine-tuned model and found that a new adversary (FGSM $\epsilon = 0.02$) can only degrade a PSMNet trained on FGSM $\epsilon=0.02$ to $13.84\%$ error and $23.74\%$ when PSMNet is trained on FGSM $\epsilon = 0.002$. We also observe these trends in DeepPruner and AANet (\figref{fig:plot-defenses}-(c, d)).

Contrary to findings reported in classification \cite{goodfellow2014explaining, kurakin2016adversarial}, augmenting the training set with adversarial examples have little negative effect on performance of stereo models for clean images. When training with $\epsilon = 0.002$ (blue), performance for PSMNet and AANet are essentially unchanged (compared to $\star$); for the largest $\epsilon = 0.02$ (orange), errors increased by $\approx$0.4\%. The smaller the norm, the less it affects performance on clean images. This is likely due to the mis-match in image intensity distributions between natural and adversarial examples. To avoid loss in performance, one can train on $\epsilon = 0.002$ and still observe the benefits on robustness. \figref{fig:plot-defenses}-(b, c, d) shows that all models are (i) robust to perturbations at $\epsilon = 0.002$ and $0.005$, (ii) comparable to leveraging larger norm perturbations when facing new adversaries, and (iii) retains original performance on clean images. 

While training on larger norms increases robustness against both existing and new adversaries, the model generally performs worse against a new adversary. For $\epsilon = 0.02$, a fine-tuned PSMNet achieves $2.74\%$ against the adversary it is trained on and $13.84\%$ against a new adversary; similarly, DeepPruner achieves $3.33\%$ and $21.39\%$ respectively. In contrast, when training on smaller norms ($\epsilon = 0.002$), the model keeps the same level robustness against existing and new adversaries. In fact, both PSMNet and DeepPruner perform better against new adversaries. For FGSM $\epsilon = 0.02$, PSMNet fine-tuned on $\epsilon = 0.002$ achieves $27.03\%$ against the existing adversary and $23.74\%$ against a new adversary; similarly, DeepPruner achieves $17.99\%$ and $13.01\%$, respectively. This phenomenon is likely because the network is overfitting to the intensity patterns of the large norm noise (also related to the slight degrade in performance). But for small norms, the network learns the underlying pattern without needing to alter its decision boundaries significantly since the intensity distribution is closer to natural images -- resulting in a more regularized model. Perhaps a strategy to learning robust stereo networks is to iteratively craft various small norm perturbations and train on them with a mixture of clean images.

\section{Discussion}
\label{sec:discussion}

Stereo networks are indeed vulnerable to adversarial perturbations. This is unexpected because the problem setting is quite unlike adversarial perturbations for single image tasks where there is no unique minimizer (a single image does not constraint the latent, the training set does). Here, the geometry of the scene can be directly observed in co-visible regions as the data term is well defined and would have a unique minimizer under mild conditions (\secref{sec:introduction}). This means that stereo matching does not require learning; learning affects regularization. So it is surprising that, despite a uniquely identifiable latent variable (disparity), the training manages to produce such a strong bias that makes the overall system susceptible to perturbations, and local perturbations to boot. What is more interesting is that, not only can these perturbations drastically alter predictions on the stereo models they are optimized for, they can also transfer across models (although with reduced potency). However, given that it is rare for a malicious agent to have full access to a network and its loss, these attacks are not feasible in practice. Yet, the fact they exist gives us an opportunity to leverage them offline and train more robust stereo networks. 

Previous works in single image based tasks have demonstrated that augmenting the training set with adversarial examples can improve robustness, but at the expense of performance on clean images. Yet, for stereo networks, we show that adversarial data augmentation can increase robustness \textit{without} compromising performance on clean images -- critical for designing robust and accurate systems. This too is likely related to the observability of the scene geometry from images where texture is sufficiently exciting. So, whereas in single image based tasks, training with adversarial perturbations alters the low-level filters to the point of hampering precision, in stereo precision is dictated by the data term, which is largely unaffected by training (correlation is an architectural inductive bias in deep stereo matching networks and is precisely why DeepPruner and AANet use it). While indeed, adversarial perturbations wreck havoc (reaching as much as 100\% in D1-all error) on networks trained only on clean images, stereo networks can recover by learning the distribution of adversarial noise through data augmentation with adversarially perturbed images and the matching process takes care of the rest.

Our work here is just a first step. We only studied transferability across models and not datasets. We also do not consider the universal setting, where a constant additive image can degrade performance across all images within a dataset. Computationally, crafting perturbations using iterative methods adds an average of $\approx$29s on top of the time needed for forward passes; hence, they cannot be computed in real time. Amongst our findings, we also observed the ``white-pixel'' phenomenon, where very little perturbations are needed to alter regions with white pixels. This is an interesting phenomenon that is present across all methods. We believe this is due to white being on the upper support of image intensities; we leave the numerical analysis of this to future work. While there is still much to do, we hope that our work can lay the foundation for harnessing adversarial perturbations to train more robust stereo models. 

\section{Ethical Impact}
\label{sec:ethical-impact}
As deep learning models are widely deployed for various tasks, adversarial examples have been treated as a threat to the security of such applications. While demonstrating that adversaries exist for stereo seems to add to this belief (since stereo is widely used in autonomous agents), we want to assure the reader that these perturbations cannot (and should not) cause harm outside of the academic setting. Cameras are not the only sensors on an autonomous agent, they are generally equipped with range sensors as well. Hence, corrupting the depth or disparity map will not cause the system to fail since it can still obtain depth information from other sources. Also, as mentioned in \secref{sec:discussion}, crafting these perturbations is computationally expensive and cannot be done in real time. Thus, we see little opportunities for negative ethical implications, but of course where there is a will there is a way.

More importantly, we see adversarial perturbations as a vehicle to develop better understanding of the behavior of black-box models. By identifying the input signals to which the output is most sensitive, we can ascertain properties of the map, as others have recently begun doing by using them to compute the curvature of the decision boundaries, and therefore the fragility of the networks and the reliability of their output.

What we want to stress in this work is that the mere existence of these perturbations tells us that there is a problem with the robustness of stereo networks. Therefore, we treat them as an opportunity to investigate and ultimately to improve stereo models. Our findings in \secref{sec:defenses} shed light on the benefits of harnessing adversarial examples and potential direction towards more robust representations.

\section{Acknowledgements}
This work was supported by ONR N00014-17-1-2072 and ARO W911NF-20-1-0158.

\bibliography{egbib}

\clearpage

\begin{center}
    {\LARGE{\textbf{Supplementary Materials}}}
\end{center}

\vspace{1em}

\begin{appendices}

\section{Summary of Contents}
\label{sec:summary-of-contents}
In \secref{sec:implementation-details}, we discuss implementation details, hyper-parameters, run-time requirements for crafting perturbations and training with adversarial data augmentation. In \secref{sec:attacks-one-image}, we show that perturbing one of the two images in a stereo pair is sufficient (though less effective) to fool stereo networks. In the main text, we demonstrated adversarial attacks, transferability, and defenses on the KITTI 2015 dataset \cite{menze2015object}; here, we repeat the experiments on the KITTI 2012 dataset \cite{geiger2012we} in \secref{sec:results-kitti-2012}. Additionally, in \secref{sec:transfer-next-frame}, we also investigate whether the adversarial perturbations crafted for a specific stereo pair is pathological by transferring them to the stereo pair taken at the next time step. Lastly, we show additional examples of perturbations crafted using each attack discussed in Sec. 3 of the main text in \figref{fig:fgsm-5e3-all}-\ref{fig:mdi2-fgsm-1e2-all}.

\begin{figure*}[ht]
    \centering
    \includegraphics[width=0.95\textwidth]{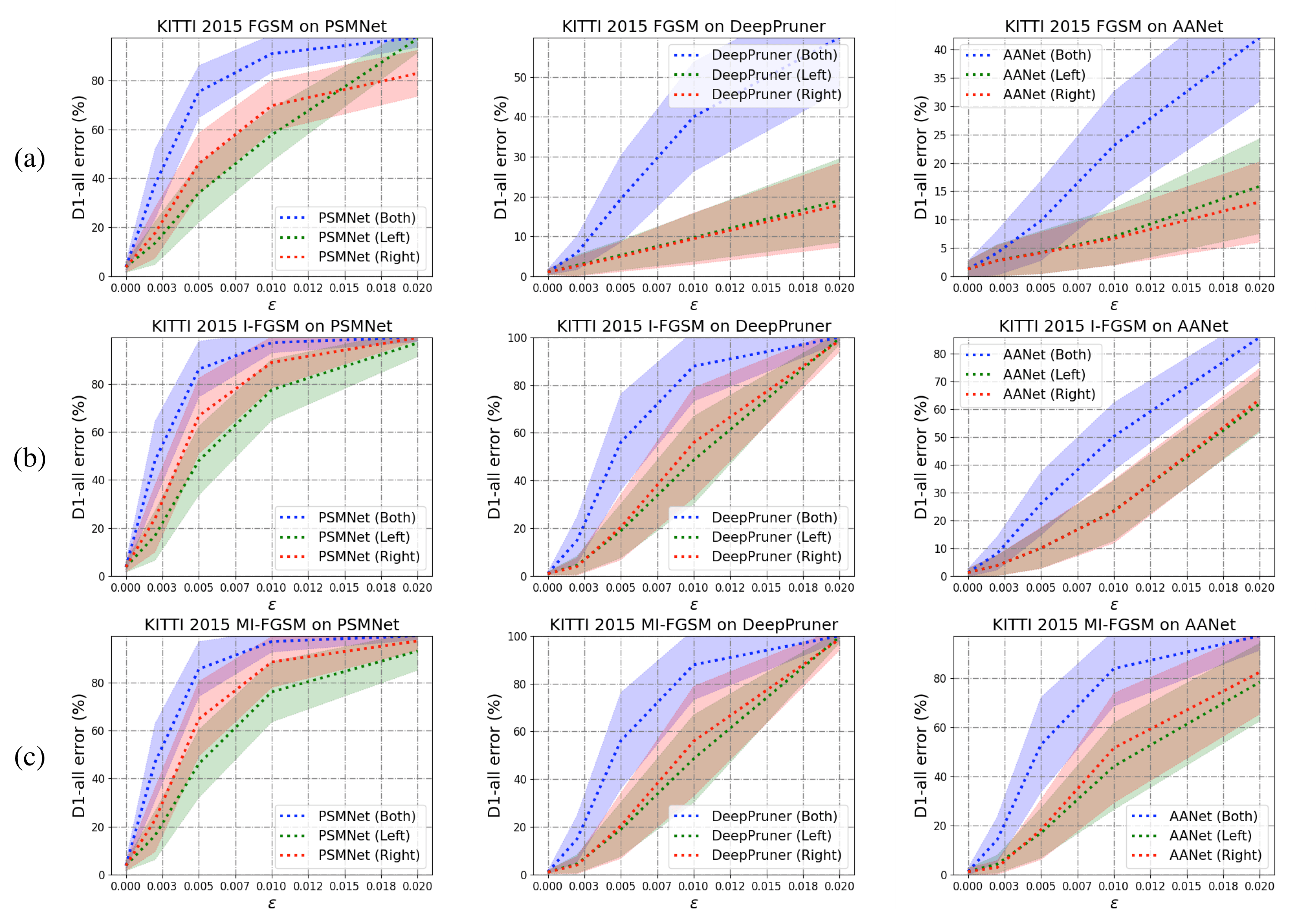}
    \caption{\textit{Comparison between attacking both images and attacking just one of the two images.} We provide quantitative comparisons on the KITTI 2015 dataset.
    Blue curve shows the error when attacking both images, green curve shows the error when attacking just the left image ($v_R = 0$) and red curve just the right image ($v_L = 0$). 
    Perturbing both images using \textbf{(a)} FGSM, \textbf{(b)} I-FGSM, and \textbf{(c)} MI-FGSM is consistently more effective than perturbing just one of them. We note that perturbing just the left image using FGSM on PSMNet yields similar results (96.98\%) at $\epsilon=0.02$. I-FGSM and MI-FGSM can also achieve over 90\% error on PSMNet and DeepPruner using $\epsilon = 0.02$.}
    \label{fig:plot-attacks-one-image}
\end{figure*}

\section{Implementation Details}
\label{sec:implementation-details}
We implemented our approach in PyTorch and used the publicly available code and pretrained models of PSMNet \cite{chang2018pyramid}, DeepPruner \cite{duggal2019deeppruner}, and AANet \cite{xu2020aanet}. We are unable to obtain the necessary computational resources to craft adversarial perturbations for each of the models on the KITTI 2012 \cite{geiger2012we} and 2015 \cite{menze2015object} datasets at full image resolution ($376 \times 1240$). Therefore, we resize the images to $256 \times 640$ resolution for PSMNet and DeepPruner, and $252 \times 636$ for AANet, when we craft adversarial perturbations. Because of the resizing, the baseline errors we report are slightly higher than those reported by the authors of PSMNet, DeepPruner and AANet. We note that PSMNet and AANet released separate pretrained models for KITTI 2012 and 2015, but DeepPruner released a single model trained on both KITTI 2012 and 2015. Therefore, for our experiments on KITTI 2012 (\secref{sec:results-kitti-2012}), we use the KITTI 2012 models for PSMNet and AANet, and the KITTI 2012, 2015 model for DeepPruner. Additionally, DeepPruner provided two pretrained models: DeepPruner-Best and DeepPruner-Fast, both trained on KITTI 2012 and 2015. We used DeepPruner-Best for all our experiments. AANet also provided two model variants: AANet and AANet+. We used AANet for all our experiments.

\begin{table}[]
    \centering
    \scriptsize
    \begin{tabular}{c c c c c}
        \toprule
            & I-FGSM & MI-FGSM & DI$^2$-FGSM & MDI$^2$-FGSM \\ \midrule
        $N$ 
            & \multicolumn{4}{c}{40} \\ \midrule
        $\alpha$, $\epsilon=0.002$
            & \multicolumn{4}{c}{0.00005} \\ \midrule
        $\alpha$, $\epsilon=0.005$
            & \multicolumn{4}{c}{0.000125} \\ \midrule
        $\alpha$, $\epsilon=0.01$
            & \multicolumn{4}{c}{0.00025} \\ \midrule
        $\alpha$, $\epsilon=0.02$
            & \multicolumn{4}{c}{0.002} \\ \midrule
        $\beta$
            & - & 0.47 & - & 0.47  \\ \midrule
        $h_{\text{min}}$
            & - & - & 0.90 & 0.90 \\ \midrule
        $h_{\text{max}}$
            & - & - & 1.00 & 1.00 \\ \midrule
        $w_{\text{min}}$
            & - & - & 0.90 & 0.90 \\ \midrule
        $w_{\text{max}}$
            & - & - & 1.00 & 1.00 \\ \midrule
        $p$
            & - & - & 0.50 & 0.50 \\ \midrule
        
    \end{tabular}
    \caption{\textit{Hyper-parameters used by iterative methods.} All iterative methods used $N = 40$. We chose $\alpha = 1/N \cdot \epsilon$ for $\epsilon~=~ \{0.01, 0.005, 0.002\}$, and $\alpha = 0.10 \epsilon$ for $\epsilon = 0.02$  for all methods. $\beta = 0.47$ is only used for momentum methods. $h_\text{min} = 0.90$, $h_\text{max} = 1.00$, $w_\text{min} = 0.90$, $w_\text{max} = 1.00$, $p = 0.50$ are only used for diversity input methods.}
    \label{tab:hyper-parameters}
\end{table}

\textbf{Hyper-parameters}. We consider perturbations for four different upper norms $\epsilon = \{0.02, 0.01, 0.005, 0.002\}$. We show the hyper-parameters used for each perturbation method in \tabref{tab:hyper-parameters}. We explored larger number of steps $N$, but found little difference. Smaller $N$ results in less effectiveness. $\epsilon = 0.002$ is where adversaries have little effect on the networks and $\epsilon = 0.02$ is the norm needed to achieve $100\%$ errors on benchmark datasets. We investigated $\alpha \in [\epsilon, \frac{1}{N}\epsilon]$ and found that larger $\alpha$ for smaller $\epsilon$ tend to be ineffective. This is likely because the accumulated perturbations quickly saturate at a small $\epsilon$ due to clipping -- additional steps nullify the effect of the perturbations. Hence we chose $\alpha = 1/N \cdot \epsilon$ for $\epsilon~=~ \{0.01, 0.005, 0.002\}$. Within the search range of $\alpha$, we found that $\alpha = 0.10$ performed the best (highest error) for $\epsilon = 0.02$. 

\textbf{Run-time}. We used an Nvidia GTX 1080Ti for our experiments. Crafting perturbations using FGSM requires on average $\approx$0.87s in addition to the time needed for a forward pass through the stereo model (PSMNet, DeepPruner, AANet). I-FGSM on average requires an extra $\approx$29.68s and MI-FGSM requires $\approx$32.23s more. DI$^2$-FGSM and MDI$^2$-FGSM on average requires $\approx$30.78s and $\approx$33.16s, respectively, in addition to the time need for a forward pass. As mentioned in the main text, while these perturbations can degrade performance and also transfer to other models, they cannot be crafted in real time. Hence, in this work we focus on leverage to learn more robust stereo networks through adversarial data augmentation.

\begin{figure*}[ht]
    \centering
    \includegraphics[width=1.00\textwidth]{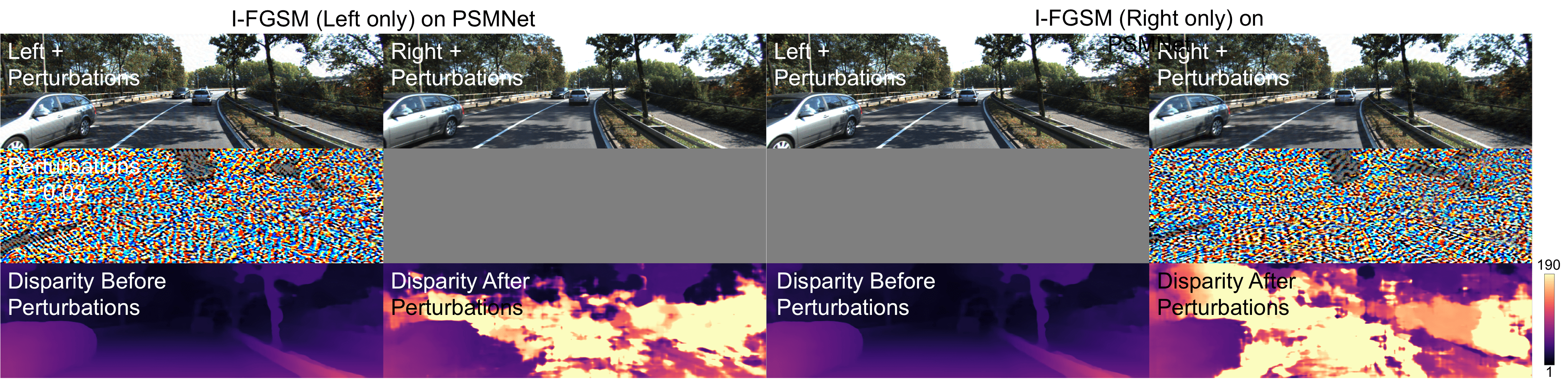}
    \caption{\textit{I-FGSM attacks on stereo networks by perturbing only the left or right image.} We demonstrate on the KITTI 2015 dataset that it is possible to fool a stereo network by only perturbing one of the two images in a stereo pair. Here, we attack PSMNet using perturbations optimized with I-FGSM ($\epsilon = 0.02$) with the constraint that it must be on either the left (where $v_R = 0$) or right (where $v_L = 0$) image. Even with perturbations on one of the two images, we can still fool the network into predicting drastically incorrect depths. Note: the shapes of salient objects (with incorrect disparities) observed in the Sec. 5, Fig. 3 of the main text are also observed here.}
    \label{fig:ifgsm-left-right-psmnet}
\end{figure*}

\begin{figure*}
    \centering
    \includegraphics[width=1.00\textwidth]{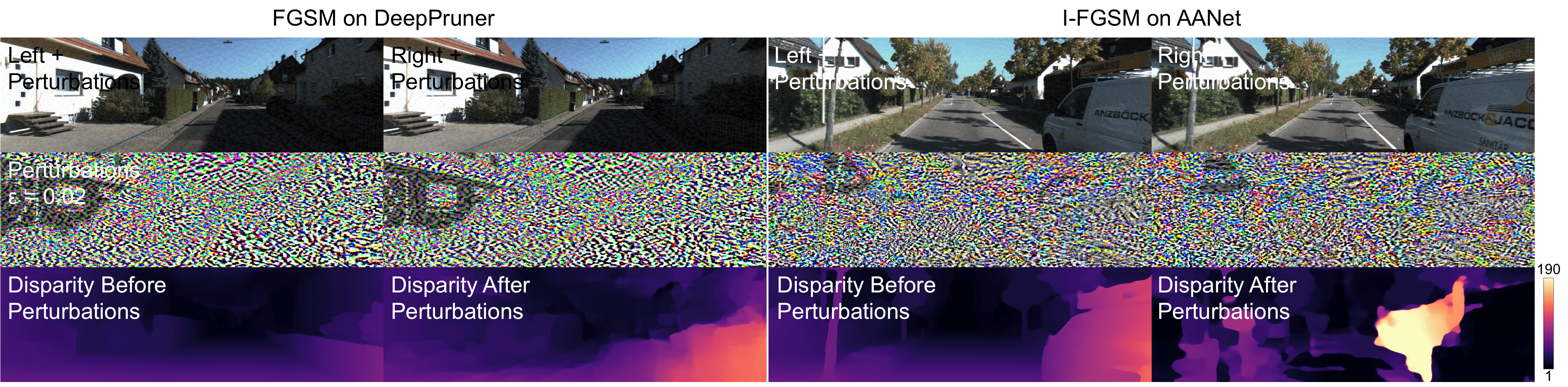}
    \caption{\textit{Qualitative examples of FGSM and I-FGSM attacks on stereo networks using stereo pairs from the KITTI 2012 dataset.} On the left, we show an FGSM attack on DeepPruner using $\epsilon = 0.02$. The perturbations mainly corrupted the right side of the output scene. On the right, we show an I-FGSM attack on AANet using $\epsilon = 0.02$. Unlike the FGSM attack, which tends to be localized, the I-FGSM attack corrupts the entire scene.}
    \label{fig:fgsm-ifgsm-kitti-2012}
\end{figure*}

\begin{figure*}[ht]
    \centering
    \includegraphics[width=0.90\textwidth]{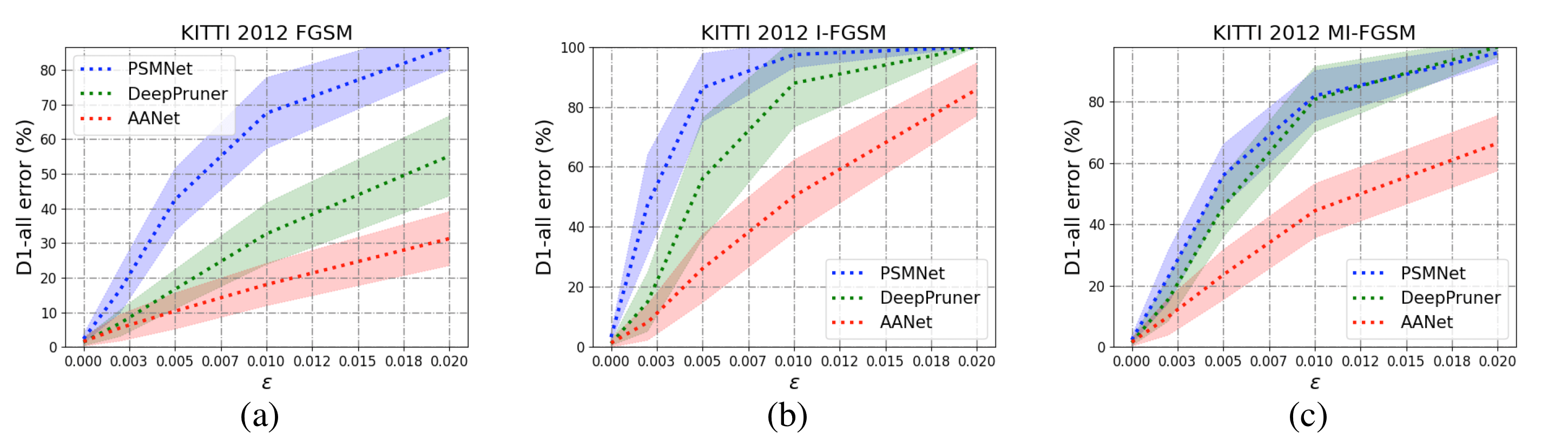}
    \caption{\textit{Attacking stereo networks.} Results of feeding stereo pairs from KITTI 2012 dataset with added adversarial perturbations crafted from \textbf{(a)} FGSM, \textbf{(b)} I-FGSM, and \textbf{(c)} MI-FGSM to PSMNet (blue), DeepPruner (green) and AANet (red). All three attacks are able to degrade the performance of all stereo methods. Iterative methods are able to achieve $\approx$100\% error for PSMNet and DeepPruner. Similar to the results demonstrated in KITTI 2015, PSMNet is the least robust, followed by DeepPruner, with AANet being the most robust.}
    \label{fig:plot-attacks-kitti-2012}
\end{figure*}

\textbf{Training}. For defending against adversaries, we fine-tuned the models on a combination of clean and perturbed images (using FGSM with various $\epsilon$). We used 160 images (and their perturbed versions) from the KITTI 2015 training set for fine-tuning, and the remaining 40 stereo pairs (and their perturbed versions) for the validation set. A similar distribution was used for experiments on KITTI 2012 dataset as well (34 stereo pairs instead of 40). All images (clean and perturbed) were resized to $256 \times 640$ resolution. PSMNet and DeepPruner took a $256 \times 512$ crop of the image during training, while AANet took a $252 \times 636$ crop of the image. We chose a learning rate of $0.001$ for PSMNet and $0.0001$ for DeepPruner and AANet after experimenting with smaller and larger learning rates. We did not use pseudo ground truth supervision in AANet during fine-tuning. We fine-tuned the models for 150 epochs.

\section{Attacking Only One Image}
\label{sec:attacks-one-image}
In Sec. 3 of the main text, we mentioned that, while we are able to fool stereo networks by introducing perturbations to both images in the stereo pair, it is also possible to fool them simply by perturbing just one of them (either the left, $v_R = 0$, or the right, $v_L = 0$, image only). \figref{fig:plot-attacks-one-image} shows a quantitative comparison between attacking both images versus attacking just one of them. Surprisingly, perturbing just one of image is sufficient to fool stereo network into predicting incorrect disparities -- although consistently with less effectiveness than perturbing both images. At the highest upper norm $\epsilon=0.02$, perturbing just the left image using FGSM is enough to degrade the performance of PSMNet to 96.98\% (note that FGSM attacks are much weaker against DeepPruner and AANet in this setting). While for smaller upper norms, all attacks on a single image are less effective than attacking both images, we note that iterative methods (I-FGSM, MI-FGSM) using $\epsilon=0.02$ can achieve similar error percentages to attacking both images for PSMNet and DeepPruner. When attacking one of the two images, trends similar to those for attacking both images are present. The larger the norm, the more effective the perturbations. AANet proves to be more robust than PSMNet and DeepPruner in this problem setting. \figref{fig:ifgsm-left-right-psmnet} shows I-FGSM attacks on PSMNet using $\epsilon=0.02$ where the perturbations are only located on either the left and right images. As we can see, the network is still fooled into predicting incorrect disparities. We note that the shapes of salient objects observed in Sec. 5, Fig. 3 of the main text are also observed here. While we see the shape of the vehicle on the left panel, the disparities of the vehicle are incorrect. This is likely due to the vehicle being co-visible and hence the shape is observed. This also shows that the network is able to learn general shapes of objects that exist in the scene. 

\begin{figure*}[ht]
    \centering
    \includegraphics[width=1.00\textwidth]{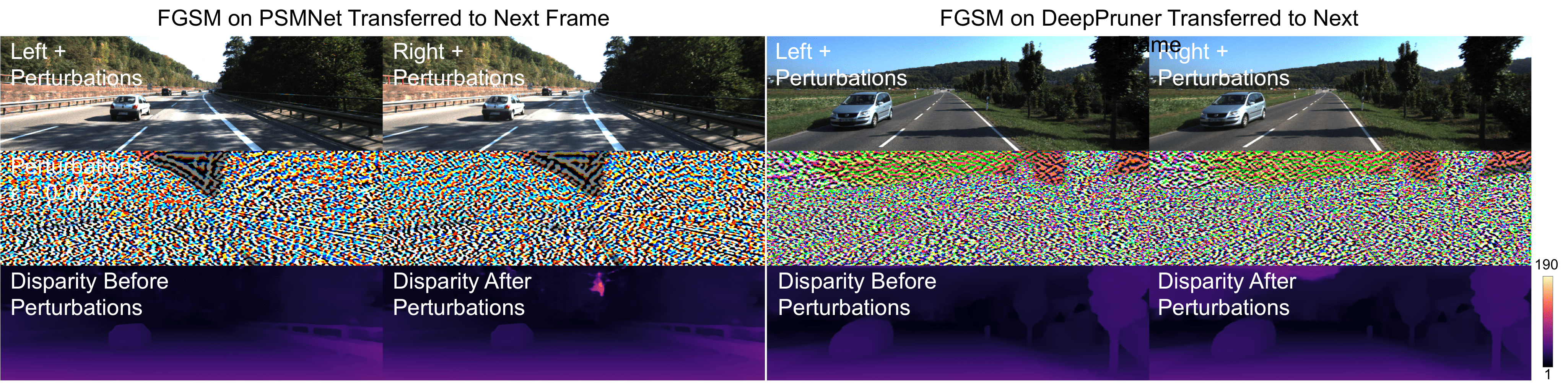}
    \caption{\textit{Investigating whether an adversarial example crafted for a specific stereo pair is pathological.} As a simple experiment, we crafted perturbations using FGSM with $\epsilon = 0.002$ for PSMNet and DeepPruner for specific stereo pairs taken at time $t$ and transferred them to the stereo pairs taken at the next time step $t+1$. Because the perturbations tend to be around the edges, we chose the smallest upper norm such that mis-match in edges between the stereo pairs taken at time $t$ and the next time step $t+1$ will not cause perturbations to be visually perceptible. We found that depending on the scene, the perturbations can transfer. For instance, if the perturbations are added onto similar structures as the previous time step, then the effect remains intact. However, if the motion is large and the intended structures are occluded or new structures are disoccluded in their place, then the perturbations will lose their effect. In the examples above, the predictions corresponding to the sky and some of the vegetation are still corrupted by the perturbations crafted for the stereo pair from the previous time step $t$.}
    \label{fig:transfer-next-frame}
\end{figure*}

\section{Results on KITTI 2012}
\label{sec:results-kitti-2012}
\begin{figure*}[ht]
    \centering
    \includegraphics[width=0.90\textwidth]{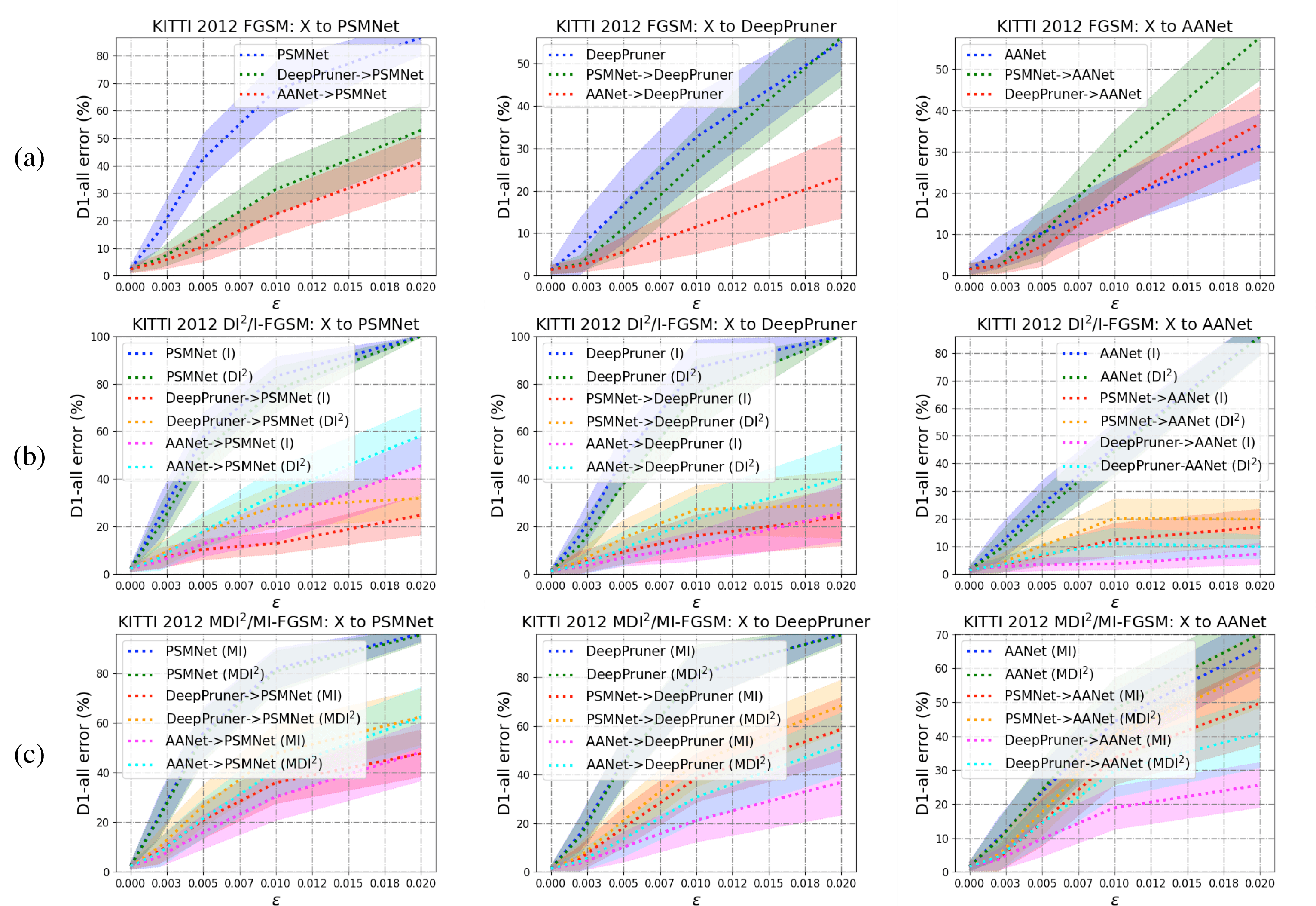}
    \caption{\textit{Transferability.}   Adversarial perturbations optimized for various models using \textbf{(a)} FGSM, \textbf{(b)} I-FGSM and DI$^2$-FGSM, and \textbf{(c)} MI-FGSM and MDI$^2$-FGSM are added to stereo pairs from the KITTI 2012 dataset and fed as input to a target model. Similar to our findings in the KITTI 2015 dataset, AANet is generally the most robust against perturbations from other models, with the exception of FGSM where perturbations optimized for PSMNet performs better those optimized for AANet. Perturbations crafted using momentum methods transfer the best. Transferability between models is, again, not symmetric. However, unlike the results in KITTI 2015, perturbations crafted for PSMNet tend to transfer the best.}
    \label{fig:plot-transferability-kitti-2012}
\end{figure*}

\begin{figure*}[h]
    \centering
    \includegraphics[width=0.90\textwidth]{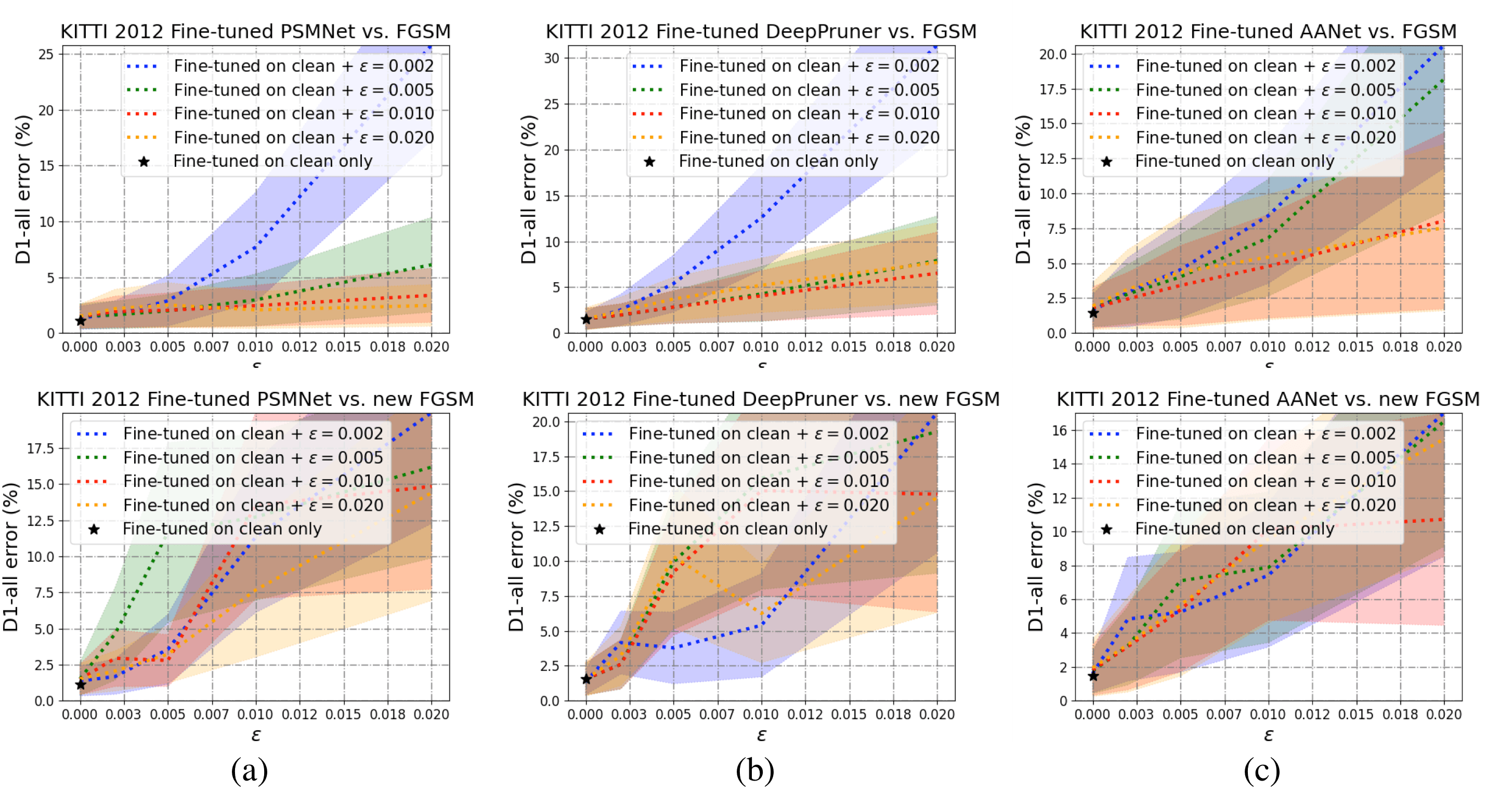}
    \caption{\textit{Training with adversarial data augmentation.} Finetuning \textbf{(a)} PSMNet, \textbf{(b)} DeepPruner and \textbf{(c)} AANet on FGSM perturbations and defending attacks against existing (top row) and new (bottom row) FGSM adversaries with various $\epsilon$. Fine-tuning on small $\epsilon$ perturbations makes the model robust against both existing and new adversaries without compromising performance on clean images. $\star$ denotes performance of training on only clean images.}
    \label{fig:plot-defenses-kitti-2012}
\end{figure*}

In the main paper, we showed experiments on the KITTI 2015 dataset. Here, we repeat all experiments performed on the KITTI 2015 dataset for KITTI 2012 dataset. \figref{fig:plot-attacks-kitti-2012} shows the effect of perturbations crafted using FGSM, I-FGSM, and MI-FGSM and \figref{fig:plot-transferability-kitti-2012} demonstrates their ability to transfer across models. \figref{fig:fgsm-ifgsm-kitti-2012} shows representative examples of FGSM and I-FGSM attacks on DeepPruner and AANet, respectively. As a defense, we fine-tuned each model on a mixture of clean and perturbed images using adversarial data augmentation. \figref{fig:plot-defenses-kitti-2012} shows the behavior of each fine-tuned model when attacked by the adversary they are trained on (top row) and also when attacked by a new adversary (bottom row).

\textbf{FGSM, I-FGSM, MI-FGSM}. For FGSM attacks (\figref{fig:plot-attacks-kitti-2012}-(a)), PSMNet proves to be the most susceptible as perturbations with $\epsilon=0.002$ can degrade performance to 16.45\% and with $\epsilon=0.02$ error increases to 86.16\%. For I-FGSM (\figref{fig:plot-attacks-kitti-2012}-(b)), perturbations can achieve 100\% error on both PSMNet and DeepPruner using the highest upper norm and 86.07\% on AANet. \figref{fig:plot-attacks-kitti-2012}-(c) shows perturbations optimized using MI-FGSM achieves 95.95\%, 97.95\%, and 66.49\% on PSMNet, DeepPruner and AANet, respectively. We show examples of FGSM and I-FGSM attacks on DeepPruner and AANet in \figref{fig:fgsm-ifgsm-kitti-2012}. In general, explicit matching methods (DeepPruner and AANet) are more robust than implicit matching or feature stacking methods (PSMNet). Overall, AANet is the most robust out of the three stereo models evaluated. 

\textbf{Transferability}. \figref{fig:plot-transferability-kitti-2012}-(a) shows the transferability of perturbations crafted by FGSM. Unlike our findings in the transferability experiments conducted on the KITTI 2015 dataset (Fig. 4, main text), PSMNet transfers the best to DeepPruner and AANet, matching performance for perturbations optimized for DeepPruner and outperforming perturbations optimized for AANet. The latter is an \textit{interesting phenomenon} also observed in the KITTI 2015 dataset -- perturbations crafted for PSMNet using FGSM are more effective than those crafted for AANet when applied to AANet. This phenomenon is isolated to only this case and we hypothesize that this is likely due to architectural similarities in the feature extraction step (general enough to DeepPruner and AANet, but not the other way around because DeepPruner and AANet features are more specific to explicit patch matching architecture). We leave additional analysis on why this occurs to future work. 

\figref{fig:plot-transferability-kitti-2012}-(b) shows I-FGSM and DI$^2$-FGSM attacks on stereo networks. Similar to our findings in KITTI 2015, I-FGSM perturbations crafted for a specific model are much less effective than FGSM when transferred to another. This is due to overfitting to the network it is optimized for. Input diversity improves transferability. Unlike Fig. 4-(b) in the main text, perturbations crafted for PSMNet transfer the best. Similar trends can be observed in \figref{fig:plot-transferability-kitti-2012}-(c). While AANet is still the most robust out of the three models, PSMNet transfers the best for KITTI 2012.

\textbf{Defenses}. To defend against adversarial attacks on stereo networks, we propose to use adversarial data augmentation and fine-tune each stereo model on a mixture of clean and adversarially perturbed (using FGSM) images. \figref{fig:plot-defenses-kitti-2012} shows the errors of each stereo method, after fine-tuning on an adversary of a specific upper norm, when attacked by adversaries optimized for the original model (top row) and by new adveraries optimized for the fine-tuned model (bottom row). As a sanity check to ensure that any performance difference is due to the adversarial data augmentation, we also fine-tuned each method on the clean data (denoted as $\star$). \figref{fig:plot-defenses-kitti-2012} shows that after training on a mixture of clean and adversarial perturbed images, all methods are now more robust against the adversary designed for the original model. Moreover, all methods are also more robust against new adversaries. Unlike findings reported in adversarial works in classification, training on adversarial examples does not compromise performance on clean images when using smaller norm ($\epsilon \in \{0.002, 0.005\}$) perturbations; when training on larger norm perturbations ($\epsilon \in \{0.01, 0.02\}$), performance only degrades slightly (performance of all methods on clean images is very close to $\star$ in \figref{fig:plot-defenses-kitti-2012}), where the change in error is $\approx$0.4\%. This is likely due to the observability of 3D from the input stereo pair e.g. one does not need to learn stereo, classic matching methods can estimate disparity without any learning. Hence, the adversarial examples serve as regularization. We note while the change in performance for larger norm is small, it is nonetheless performance degradation; we hypothesize that this is due to the mis-match in intensity distribution between the clean and perturbed images.

\section{Transferring Perturbations to Next Frame}
\label{sec:transfer-next-frame}
To investigate whether or not the adversarial perturbations crafted for a specific stereo pair is pathological -- in that they can only do damage to the stereo pair that they are optimized for, we perform a simple experiment of transferring (adding) the perturbations crafted for a stereo pair at time step $t$ to the temporally adjacent stereo pair at time step $t+1$. Because groundtruth disparity is not available for stereo pairs taken at time $t+1$, we only evaluate this section qualitatively in \figref{fig:transfer-next-frame}. \figref{fig:transfer-next-frame} shows examples of stereo pairs at time $t+1$ with added perturbations crafted the stereo pair at time $t$, and the predictions for the images taken at $t+1$ -- before and after adding the perturbations.

\figref{fig:transfer-next-frame} shows that the perturbations have the ability to transfer to the frames taken at the next time step; however, it depends on the structures in the scene. For this experiment, we crafted FGSM perturbations for PSMNet, DeepPruner and AANet using the smallest norm $\epsilon = 0.002$. Because the perturbations tend to concentrate around edges, we chose the smallest upper norm such that mis-match in edges between the stereo pair at time $t$ and the next time step $t+1$ will not cause perturbations to be visually perceptible. We found that depending on the scene, the perturbations can transfer. For instance, if the perturbations are added onto similar structures that are present in the previous time step, then the effect remains intact. However, if the motion is large and the intended structures are occluded or new structures are disoccluded, then it loses the effect. \figref{fig:transfer-next-frame} demonstrates this phenomenon. The predictions for the sky and some of the vegetation regions are still corrupted by the perturbations crafted for the stereo pair from the previous time step. Based on our observations, we do not expect the perturbations optimized for one stereo pair to retain the same effect on a stereo network when added to a different stereo pair of a different scene.

\section{Additional Examples}
\label{sec:additional-examples}
\begin{figure*}[ht]
    \centering
    \includegraphics[width=0.95\textwidth]{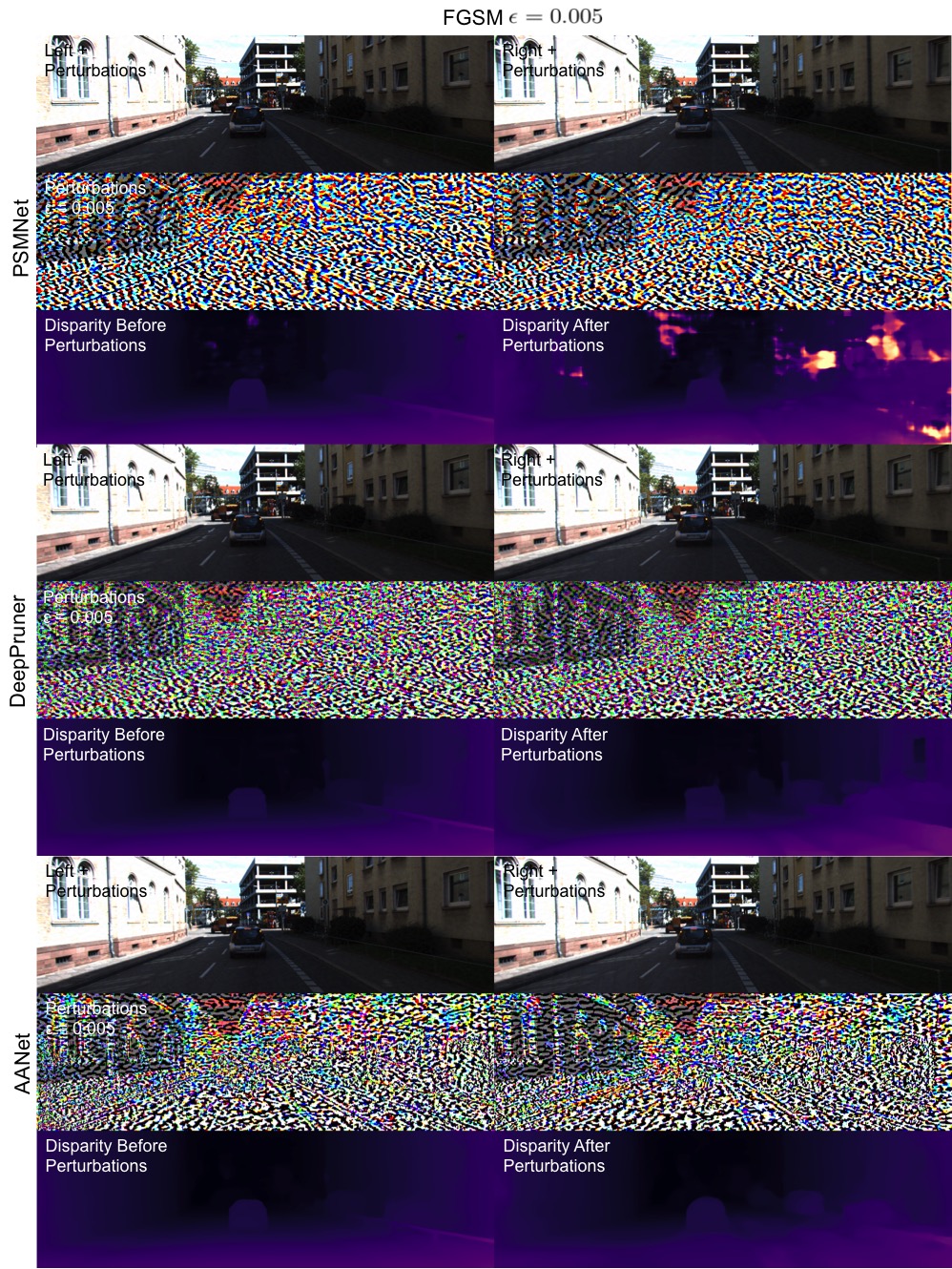}
    \caption{\textit{Examples of FGSM attacks on PSMNet, DeepPruner and AANet using $\epsilon = 0.005$.}}
    \label{fig:fgsm-5e3-all}
\end{figure*}

\begin{figure*}[ht]
    \centering
    \includegraphics[width=0.95\textwidth]{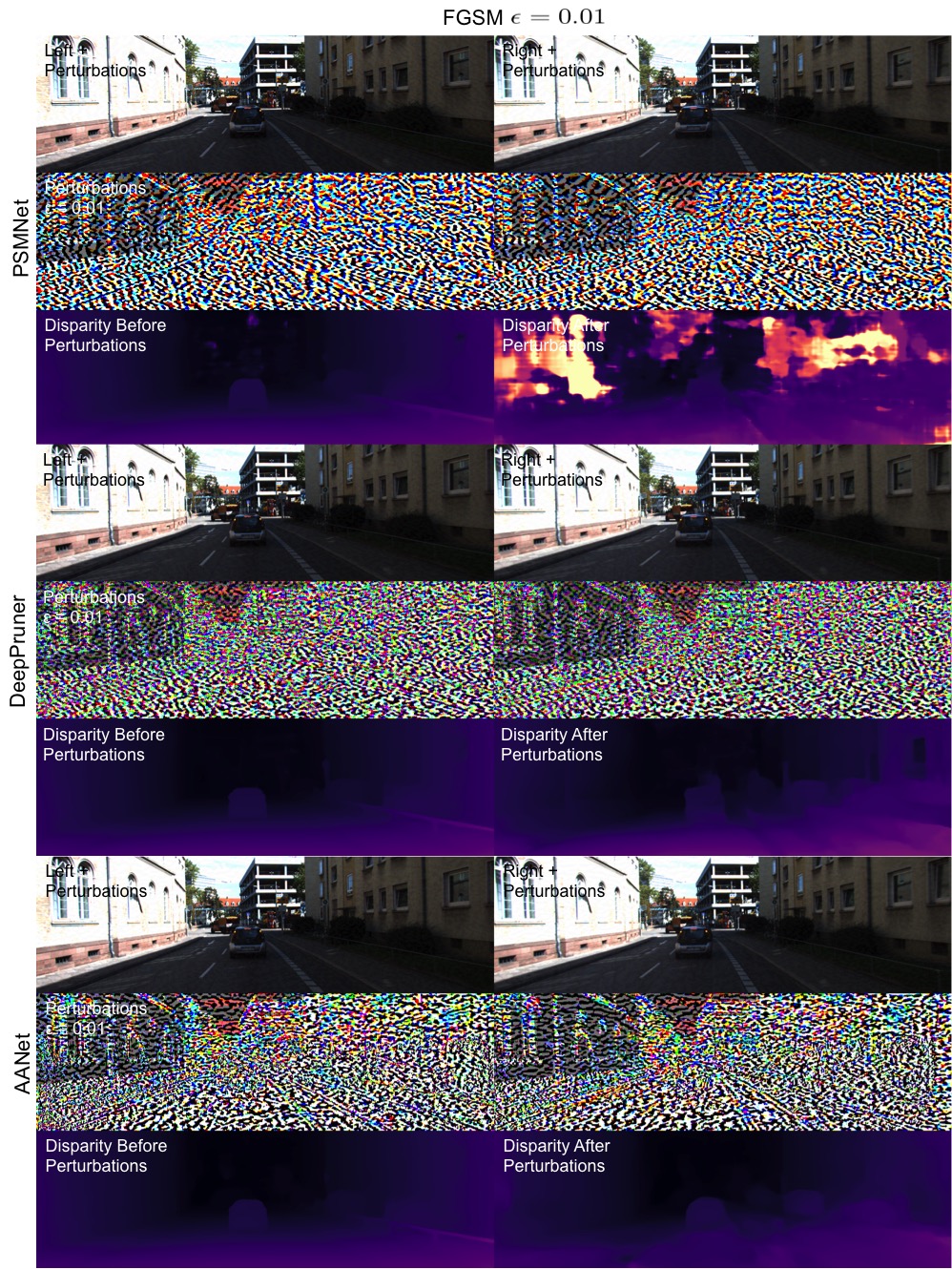}
    \caption{\textit{Examples of FGSM attacks on PSMNet, DeepPruner and AANet using $\epsilon = 0.01$.}}
    \label{fig:fgsm-1e2-all}
\end{figure*}

\begin{figure*}[ht]
    \centering
    \includegraphics[width=0.95\textwidth]{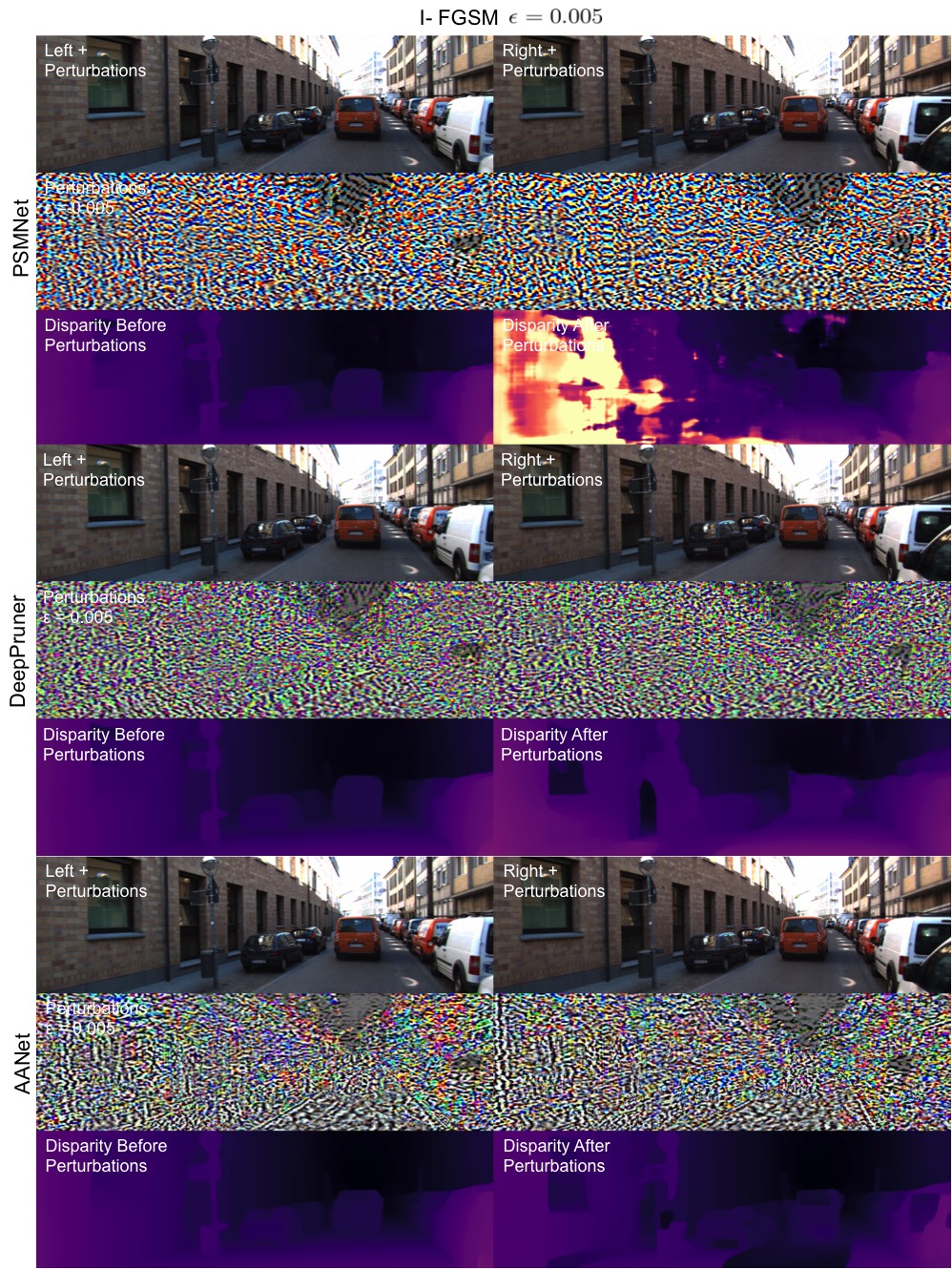}
    \caption{\textit{Examples of I-FGSM attacks on PSMNet, DeepPruner and AANet using $\epsilon = 0.005$.}}
    \label{fig:i-fgsm-5e3-all}
\end{figure*}

\begin{figure*}[ht]
    \centering
    \includegraphics[width=0.95\textwidth]{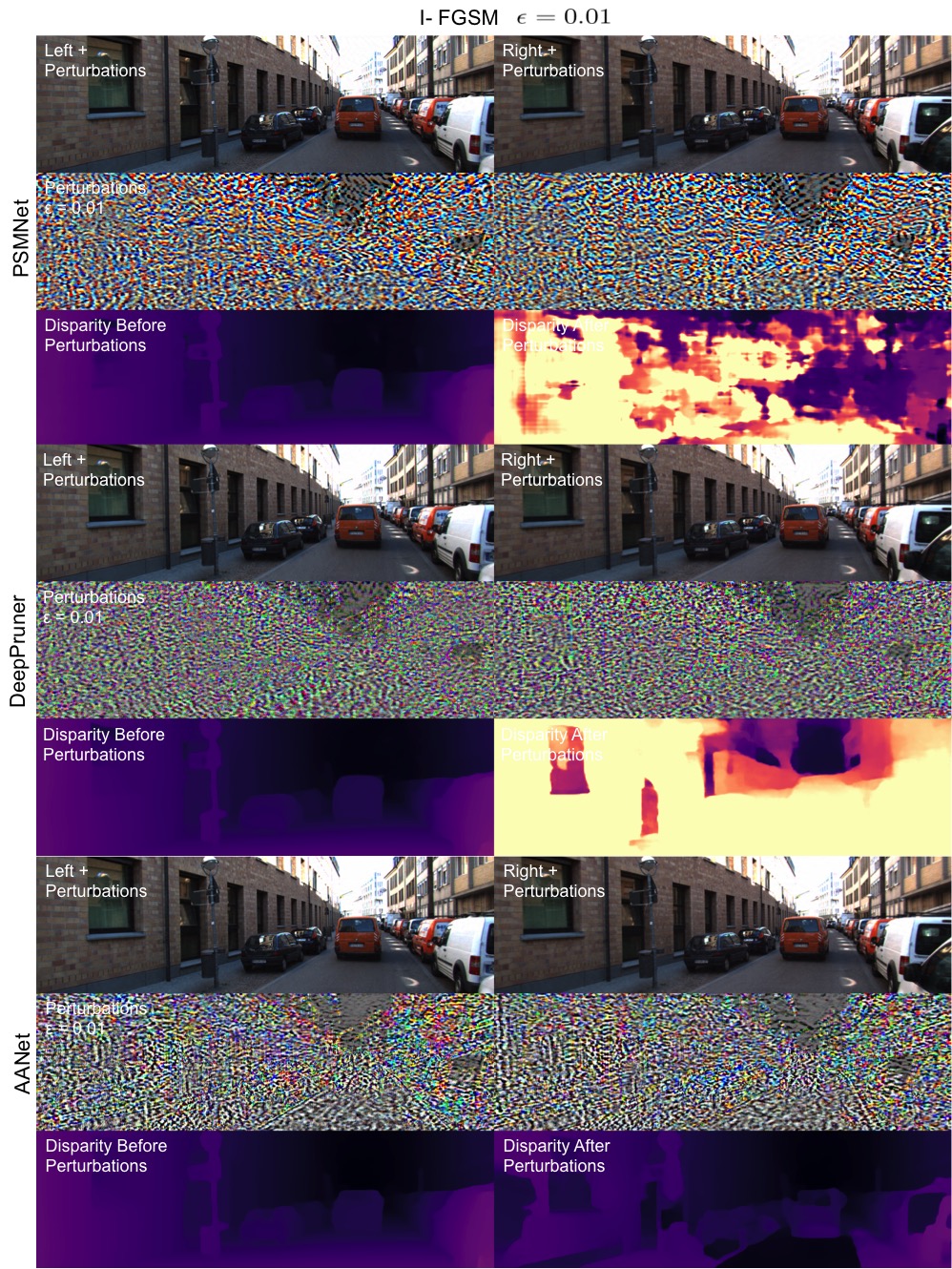}
    \caption{\textit{Examples of I-FGSM attacks on PSMNet, DeepPruner and AANet using $\epsilon = 0.01$.}}
    \label{fig:i-fgsm-1e2-all}
\end{figure*}

\begin{figure*}[ht]
    \centering
    \includegraphics[width=0.95\textwidth]{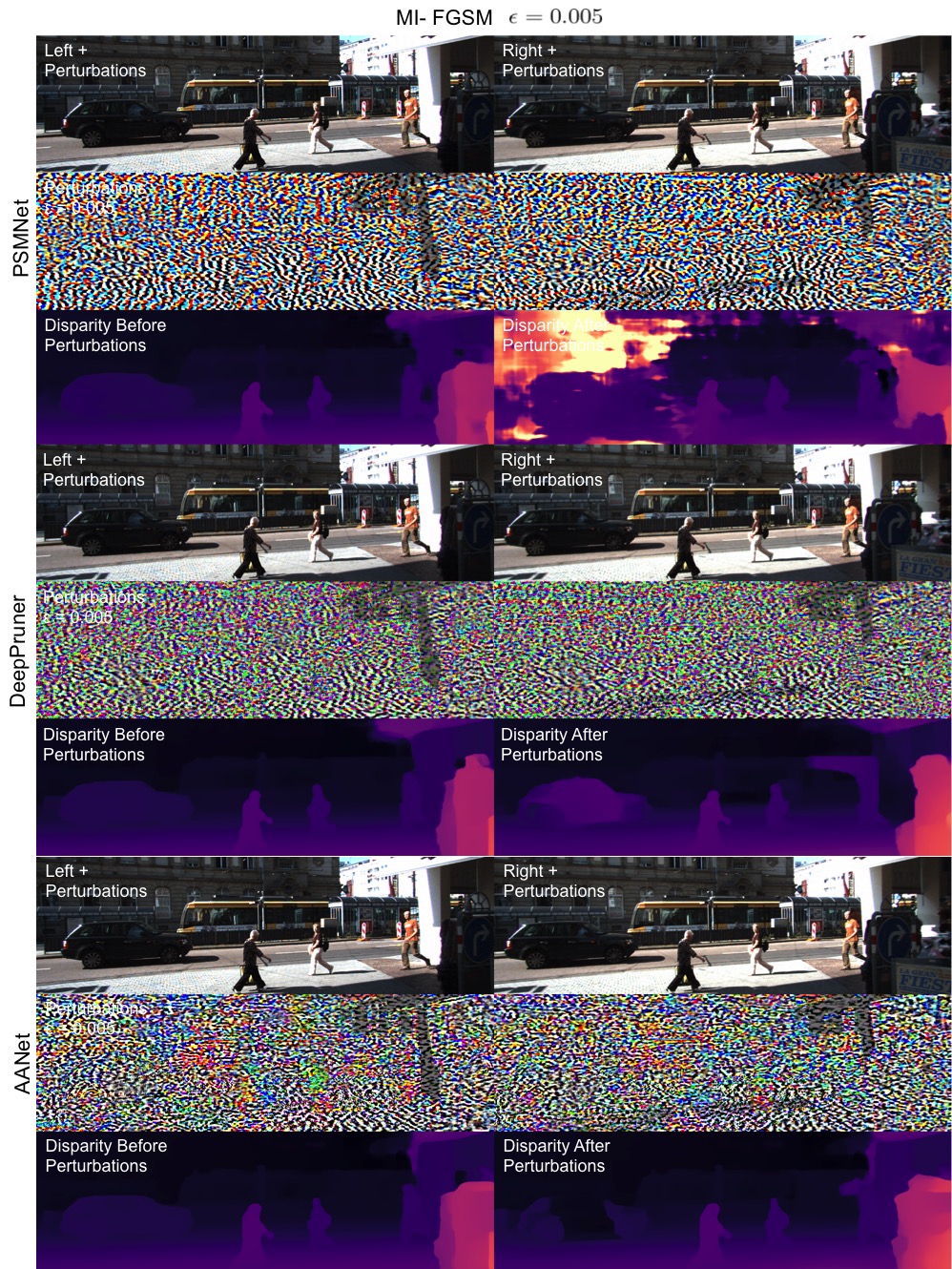}
    \caption{\textit{Examples of MI-FGSM attacks on PSMNet, DeepPruner and AANet using $\epsilon = 0.005$.}}
    \label{fig:mi-fgsm-5e3-all}
\end{figure*}

\begin{figure*}[ht]
    \centering
    \includegraphics[width=0.95\textwidth]{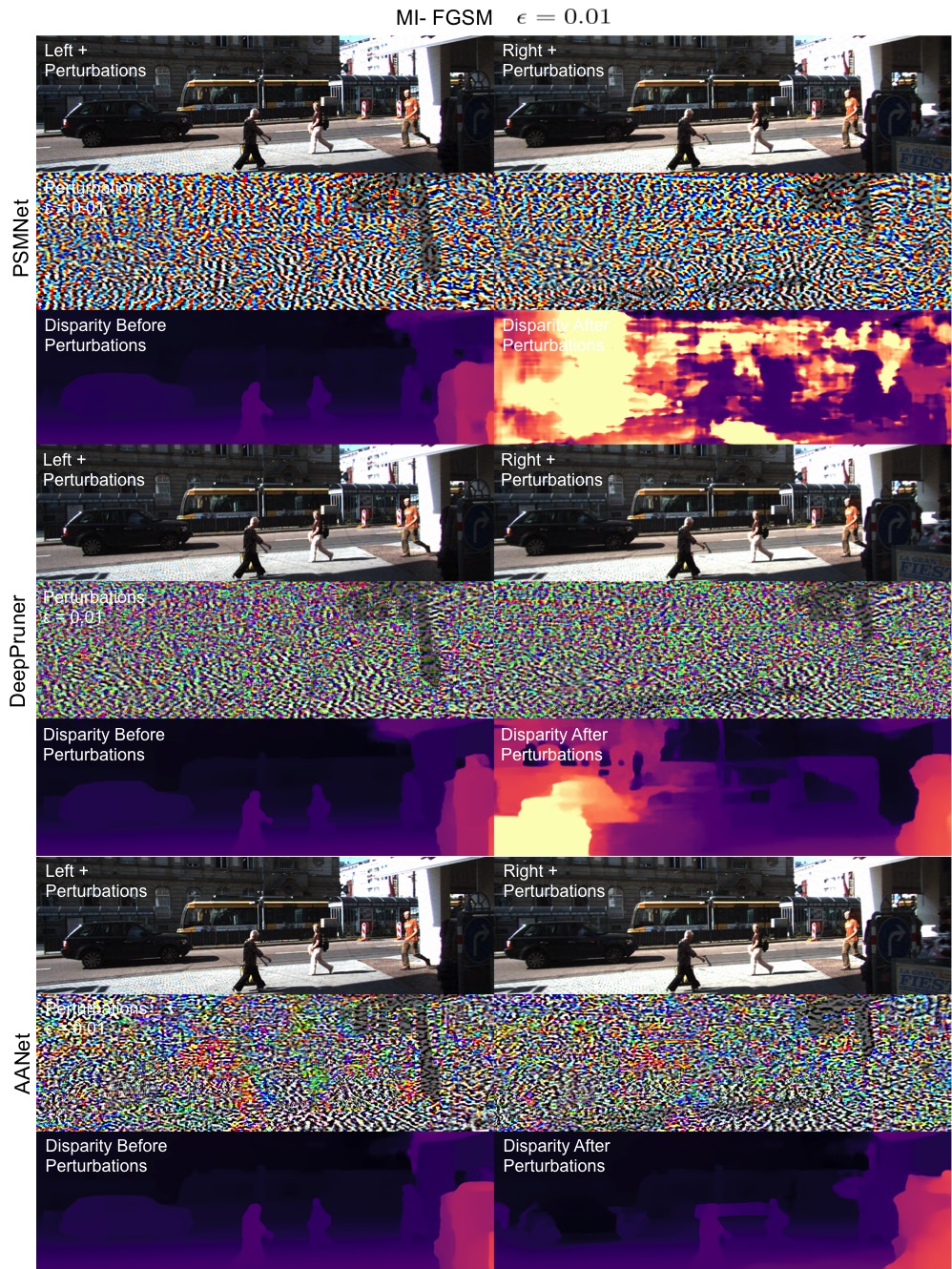}
    \caption{\textit{Examples of MI-FGSM attacks on PSMNet, DeepPruner and AANet using $\epsilon = 0.01$.}}
    \label{fig:mi-fgsm-1e2-all}
\end{figure*}

\begin{figure*}[ht]
    \centering
    \includegraphics[width=0.95\textwidth]{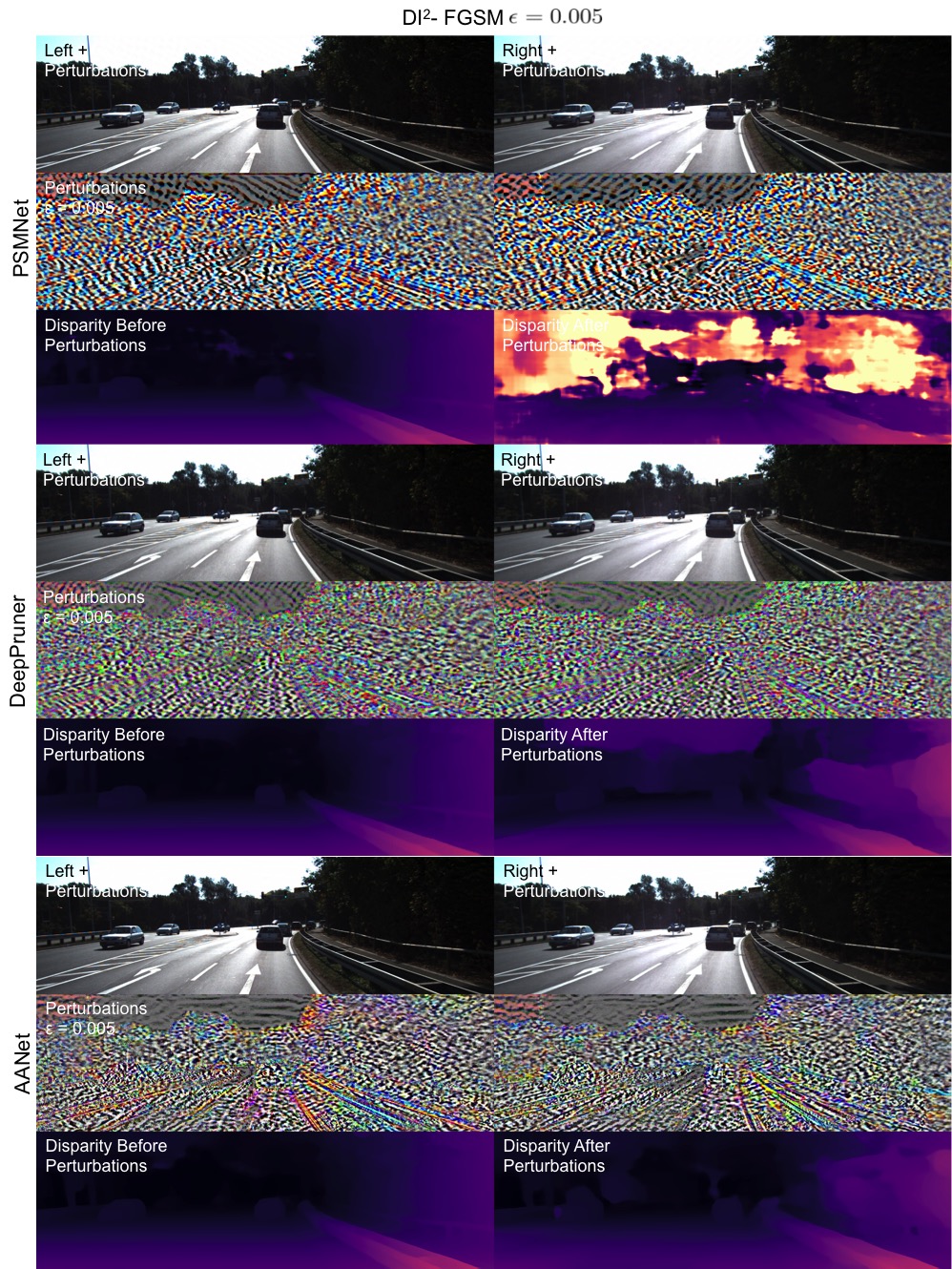}
    \caption{\textit{Examples of DI$^2$-FGSM attacks on PSMNet, DeepPruner and AANet using $\epsilon = 0.005$.}}
    \label{fig:di2-fgsm-5e3-all}
\end{figure*}

\begin{figure*}[ht]
    \centering
    \includegraphics[width=0.95\textwidth]{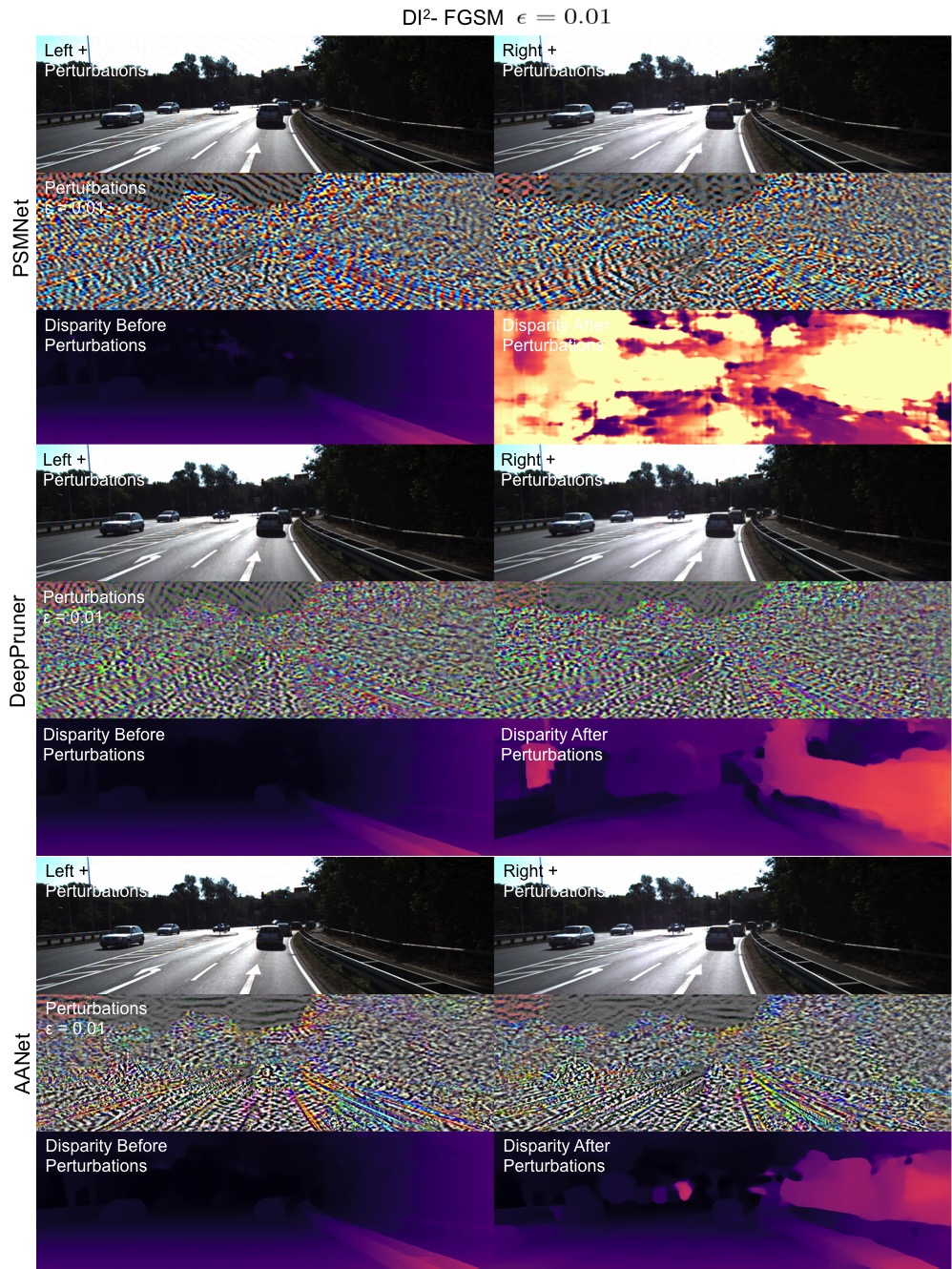}
    \caption{\textit{Examples of DI$^2$-FGSM attacks on PSMNet, DeepPruner and AANet using $\epsilon = 0.01$.}}
    \label{fig:di2-fgsm-1e2-all}
\end{figure*}

\begin{figure*}[ht]
    \centering
    \includegraphics[width=0.95\textwidth]{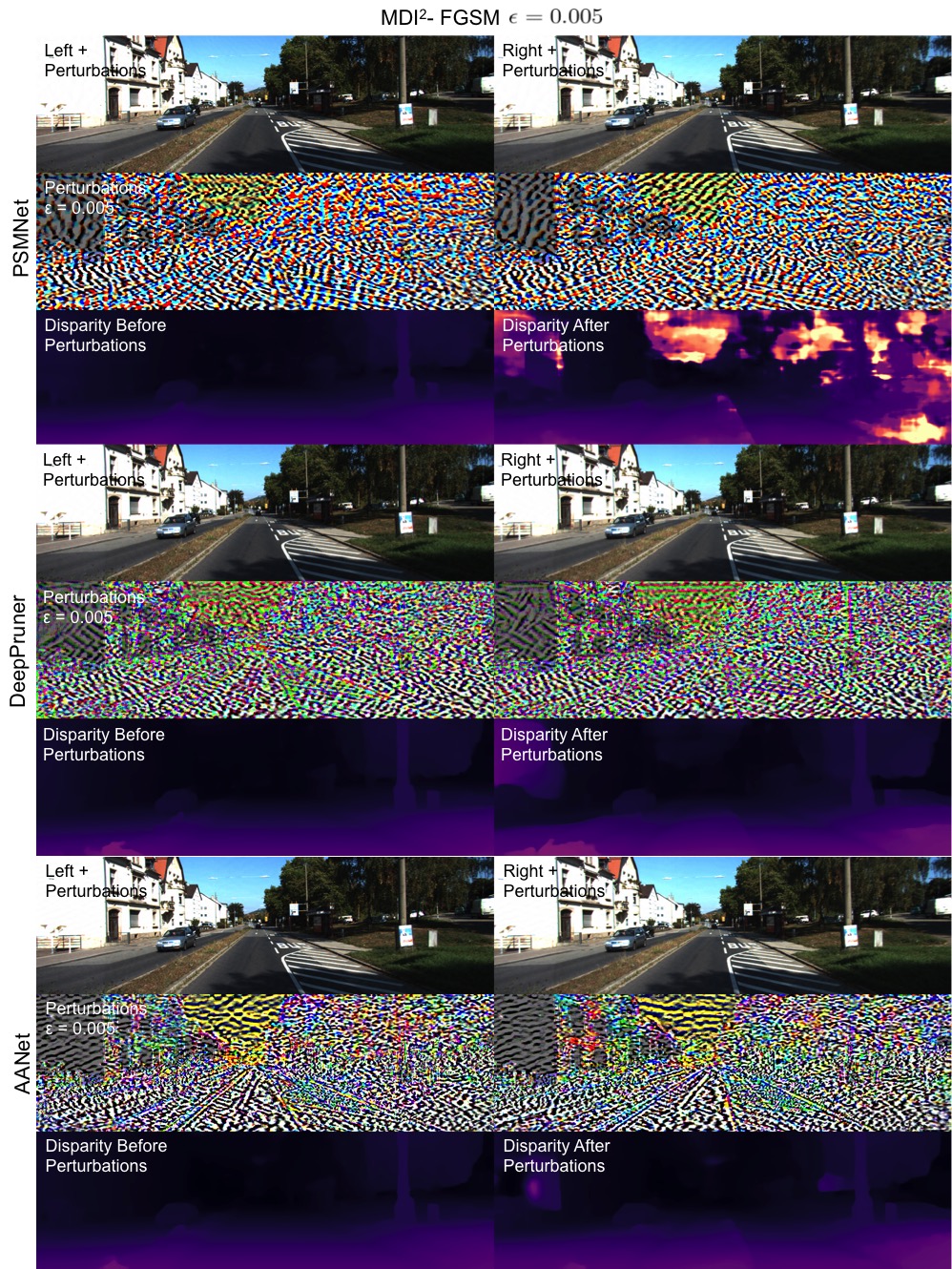}
    \caption{\textit{Examples of MDI$^2$-FGSM attacks on PSMNet, DeepPruner and AANet using $\epsilon = 0.005$.}}
    \label{fig:mdi2-fgsm-5e3-all}
\end{figure*}

\begin{figure*}[ht]
    \centering
    \includegraphics[width=0.95\textwidth]{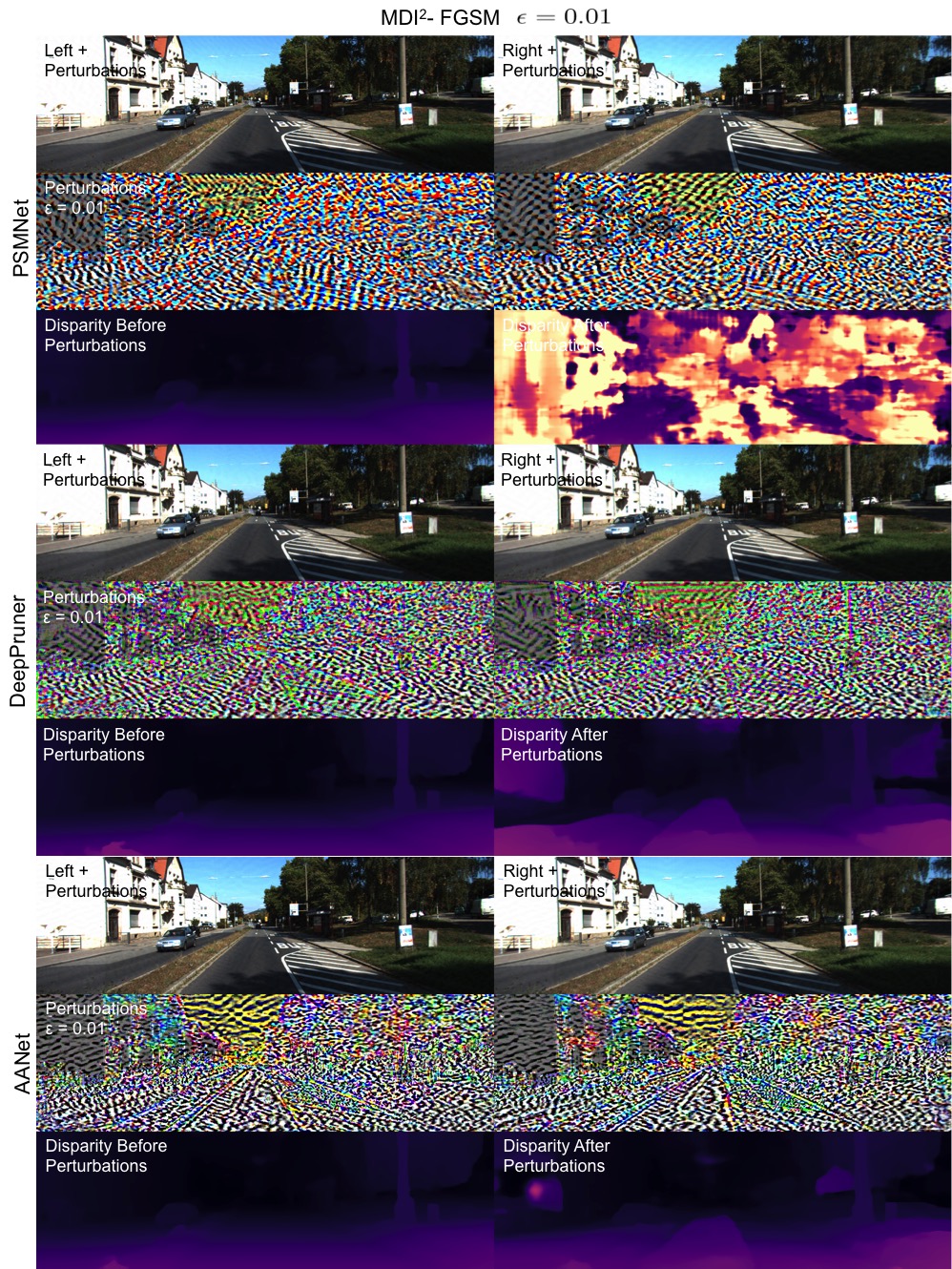}
    \caption{\textit{Examples of MDI$^2$-FGSM attacks on PSMNet, DeepPruner and AANet using $\epsilon = 0.01$.}}
    \label{fig:mdi2-fgsm-1e2-all}
\end{figure*}

In the main text, we have primarily shown examples of perturbations with upper norms $\epsilon \in \{0.002, 0.02\}$ (smallest and largest norms considered) crafted using various methods. \figref{fig:fgsm-5e3-all} to \ref{fig:mdi2-fgsm-1e2-all} show examples of attacks on PSMNet, DeepPruner, and AANet using $\epsilon \in \{0.005, 0.01\}$. \figref{fig:fgsm-5e3-all} and \ref{fig:i-fgsm-1e2-all} compare FGSM attacks for the two norms on the same stereo pairs, \figref{fig:fgsm-5e3-all} and \ref{fig:i-fgsm-1e2-all} compare I-FGSM attacks, \figref{fig:mi-fgsm-5e3-all} and \ref{fig:mi-fgsm-1e2-all} compare MI-FGSM attacks, \figref{fig:di2-fgsm-5e3-all} and \ref{fig:di2-fgsm-1e2-all} compare DI$^2$-FGSM attacks and lastly, \figref{fig:mdi2-fgsm-5e3-all} and \ref{fig:mdi2-fgsm-1e2-all} compare MDI$^2$-FGSM attacks. Increasing the upper norm from 0.005 to 0.01 does not make the perturbations visible; however, it does increase the effect of perturbations to fool the stereo networks into predicting drastically different scenes.

\end{appendices}

\end{document}